\documentclass[runningheads]{llncs}
\usepackage{makeidx}
\usepackage{graphicx}
\usepackage{amsmath,amssymb} 
\usepackage{booktabs}
\usepackage{color}
\usepackage{rotating}

\newcommand{\myparagraph}[1]{\noindent\textbf{#1}}

\newcommand{\mysubparagraph}[1]{\noindent\textit{#1}}

\newcommand{\tabincell}[2]{\begin{tabular}{@{}#1@{}}#2\end{tabular}}

\begin{document}
	\pagestyle{headings}
	\mainmatter


	\title{Learning Dilation Factors for\\Semantic Segmentation of Street Scenes}

	\titlerunning{Learning Dilation Factors for Semantic Segmentation of Street Scenes}
	\authorrunning{Yang He, Margret Keuper, Bernt Schiele, Mario Fritz}
	\author{Yang He$^1$, Margret Keuper$^2$, Bernt Schiele$^1$, Mario Fritz$^1$}
	\institute{$^1$ Max Planck Institute for Informatics, Saarbr\"ucken, Germany \\ $^2$University of Mannheim, Mannheim, Germany}

	\maketitle

	\begin{abstract}
	Contextual information is crucial for semantic segmentation. However, finding the optimal trade-off between keeping desired fine details and at the same time providing sufficiently large receptive fields is non trivial. This is even more so, when objects or classes present in an image significantly vary in size. Dilated convolutions have proven valuable for semantic segmentation, because they allow to increase the size of the receptive field without sacrificing image resolution. However, in current state-of-the-art methods, dilation parameters are hand-tuned and fixed. In this paper, we present an approach for learning dilation parameters adaptively per channel, consistently improving semantic segmentation results on street-scene datasets like Cityscapes and Camvid.
	\end{abstract}

	\section{Introduction}
	\label{sec:introduction}
	Semantic segmentation is the task of predicting the semantic category for each pixel in an image, i.e. its class label from a given set of labels. It is considered a crucial step towards scene understanding and has a wide range of use-cases including autonomous driving and service robotics. 
The trade-off between local detail and global context is inherent in the task. The prediction of  class labels requires sufficient contextual information, especially for semantic classes whose instances usually cover large portions of the image (e.g. {\it trucks}, {\it street}) or may lack local features (e.g. {\it sky}). At the same time, well localized detailed information is important for pixel-accurate prediction. 

Recently, dilated convolutions have been proposed to improve semantic segmentation performance by providing larger receptive fields without sacrificing image resolution or adding network complexity \cite{YuKoltun2016}. While this principled idea has shown promise for recent architectures \cite{pspnet2016zhao,chen2016deeplab}, the dilation parameters are not learned but hand-tuned and fixed. In contrast, we propose to learn the dilation parameters end-to-end thus
%
 generalizing the concept of dilated convolutions \cite{YuKoltun2016}. More specifically, we propose a fully trainable dilated convolution layer that allows to not only learn dilation parameters for each convolutional layer but for each channel individually. Thus, different features can be extracted and combined at different scales, rendering the network more flexible with respect to its receptive fields. 
We leverage the proposed layer to facilitate the learning of dilation parameters within three different network architectures for semantic segmentation, specifically  Deeplab-LargeFOV \cite{chen2016deeplab}, Deeplab-v2 \cite{chen2016deeplab} and PSPNet \cite{pspnet2016zhao}. For the task of street scene segmentation, we show that the proposed method consistently improves results over the respective baselines.


	\myparagraph{Related Work:}
	\label{sec:related}
	Several large-scale datasets have been released  recently to research semantic segmentation. Autonomous driving scenarios are prominently represented for example with the Camvid \cite{camvid2008} and Cityscapes \cite{cordts2016cityscapes} benchmark datasets. In this work, we focus on semantic segmentation in such street views, and build our model on recent successful architectures in this domain.

\mysubparagraph{Dilated convolutions for semantic segmentation.}
Contextual information plays a crucial role in semantic segmentation. Dilated convolutions \cite{YuKoltun2016} were introduced to aggregate context information without necessarily downsampling the resolution. 
 With a defined dilation factor, convolution operations are performed by sampling discrete locations. 
Dilated convolution architectures are widely used in recent semantic segmentation methods for their ability to capture large context while preserving fine details.
Chen \emph{et al.} \cite{chen2016deeplab} 
propose to utilize a large dilation factor in the original Deeplab model \cite{chen2014semantic} to provide large context, leading to better performance. 
However, local information is beneficial to recognize fine details. To leverage local and wide context information, Chen \emph{et al.} \cite{chen2016attention} present the Deeplab-v2 model with atrous spatial pyramid pooling (ASPP), which combines multi-level dilated convolutions and report improved performance.

\mysubparagraph{State-of-the-art architectures for street scenes.}
Recent state-of-the-arts are built on deep residual networks (ResNets) \cite{he2016deep}, which are powerful in various vision tasks. Pohlen \emph{et al.} \cite{pohlen2017full} propose a full-resolution residual network, which contains a two-stream architecture and achieves state-of-the-art performance on  public benchmarks \cite{cordts2016cityscapes}. One stream performs on the full image resolution, capturing image details and precise image boundaries. The second stream undergoes a sequence of pooling operations to extract visual features for robust recognition.
Zhao \emph{et al.} \cite{pspnet2016zhao} propose the pyramid scene parsing network (PSPNet), which provides multi-level global context information and achieve good performance in street view semantic segmentation.

	\myparagraph{Contributions:}
	\label{sec:contribution}
	While dilation parameters for dilated convolutions are fixed and manually-tuned in previous work
\cite{pspnet2016zhao,chen2016deeplab,YuKoltun2016}, we provide a method allowing to learn those parameters for each channel individually.
Overall, the main contributions of this paper are:
%
(1) We propose a learnable channel-based dilated convolution layer, whose dilation factors can be fractional numbers. The proposed layer is compatible with state-of-the-art CNN architectures and can be trained end-to-end without extra supervision;
(2) We improve the recent state-of-the-art semantic segmentation networks PSPNet \cite{pspnet2016zhao} and Deeplab-v2 \cite{chen2016deeplab} in various street view datasets \cite{cordts2016cityscapes,camvid2008,geiger2012we} by replacing fixed dilated convolutions with the proposed layer. Our method achieves consistent improvement over baselines and yields visually more convincing predictions.
    
	\section{Learning Dilation Factors for Convolutions}
    \label{sec:method}
\begin{figure}[!t]
\begin{center}
\begin{tabular}{ccc}
\multicolumn{3}{c}{
\includegraphics[width=0.8\linewidth]{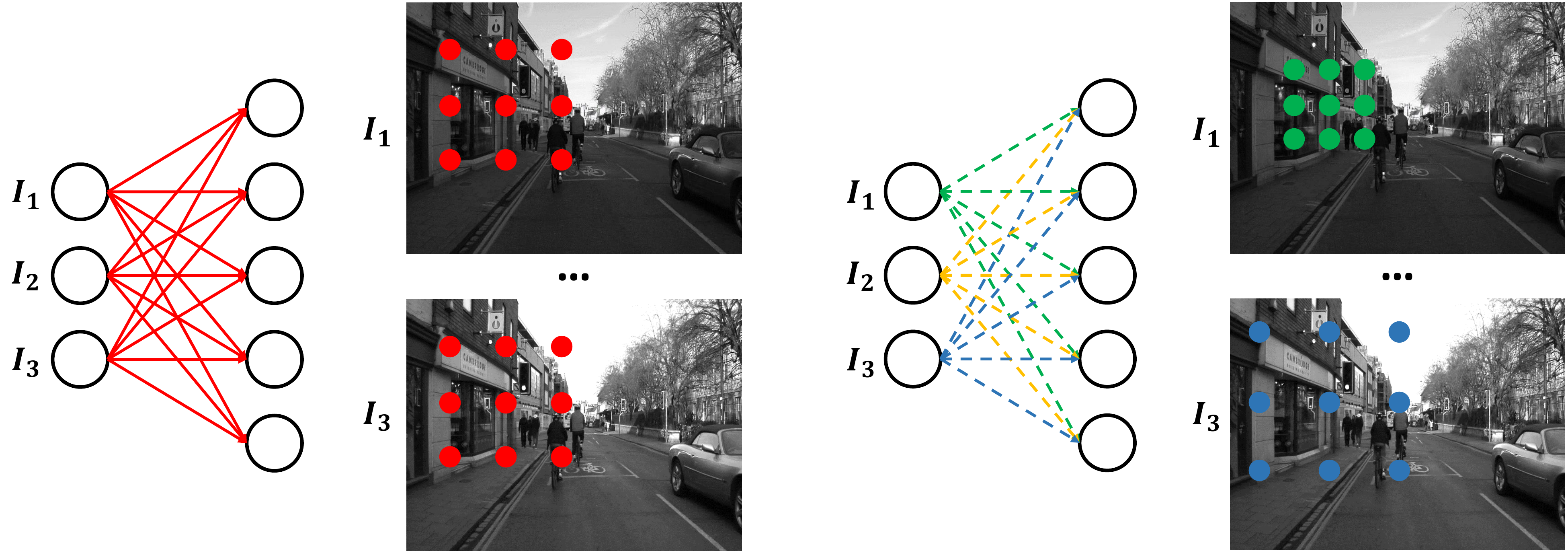}}\\
   \phantom{dilated} Dilated convolutions \cite{YuKoltun2016} &\phantom{dilated }&
   Learnable dilated convolutions
   \end{tabular}
   \label{fig:proposed_dilation}
\end{center}
\caption{Illustration of standard dilated convolutions (left) and the proposed channel based dilated convolutions (right). Standard dilated convolutions have a constant, manually set (solid lines) and integer valued dilation parameter for different channels. The proposed layer allows for channel-wise learning (dash lines) of dilation factors (encoded with different colors), which can take fractional values.}
\end{figure}

In this section, we describe the concept and formulation of the 
channel based fractional dilated convolution layer as illustrated in Fig.~1.
Our proposed layer is a generalization of dilated convolutions as they were proposed in \cite{YuKoltun2016} (compare Fig.~1 (left)). Dilated convolutions provide a simple module facilitating to aggregate context information without pooling or downsampling the original image. It thus allows to preserve high spatial resolution. 
While previous dilated convolutions 
require manual tuning of an integer valued dilation parameter, the  proposed method facilitates to learn dilation parameters from training data via back propagation.
Thus, to allow for the definition of a gradient on the dilation parameter, these parameters can no longer be constrained to integer values but are relaxed to take a value in $\mathbb{R}^+$. 
To add further flexibility to the network w.r.t. the amount of context provided to each layer and each channel, we further allow for a channel-wise optimization of dilation parameters (compare Fig.~1 (right)).

For a learned, fractional dilation factor, the output feature map of the dilated convolutions is computed using bilinear interpolation -- inspired by spatial transformer networks \cite{jaderberg2015spatial}.
The proposed learnt dilated convolution layer is compatible with existing architectures, as it generalizes (and therefore can replace) convolutional and dilated convolutional layers in a given network architecture. 

\subsection{Forward pass}
\label{subsec:forward}
We first give a brief recap on conventional dilated convolutions. With filter weights $\textbf{W}$ and a bias term $\textbf{b}$, the input feature $\textbf{X}$ can be transformed to the output feature $\textbf{Y}$ by
\begin{equation}
\begin{split}
\textbf{Y}  &=  \textbf{W}\ast \textbf{X} + \textbf{b},\qquad or\\
y_{m,n}  &=  \sum_{c}\sum_{i,j} w_{c,i,j}\cdot x_{c,m+i\cdot d,n+j\cdot d} + b,\\
\end{split}
\label{eq:convention_convolution}
\end{equation}
where $d$ is the dilation factor, and must be an integer.
\begin{figure}[!t]
\begin{center}
   \includegraphics[width=0.7\linewidth]{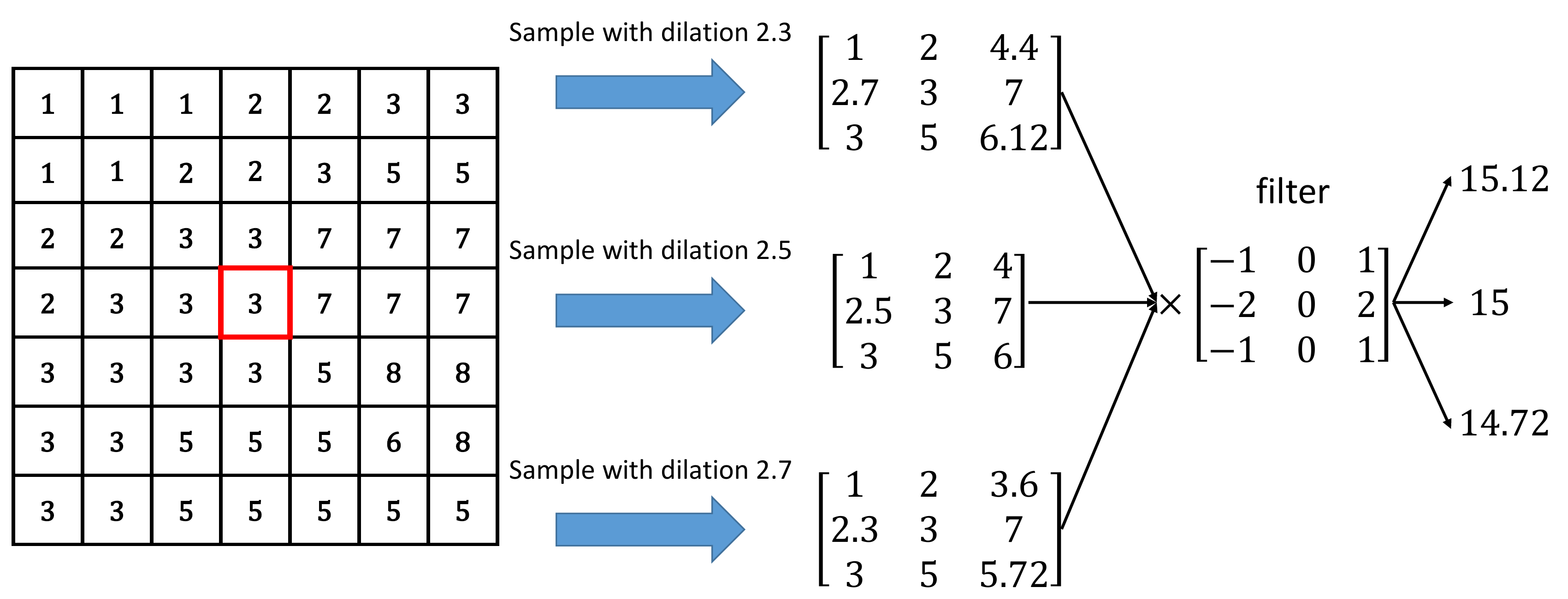}
   \label{fig:bilinear_interpolation}
\end{center}
 \caption{An example of the proposed dilated convolutions with a fractional dilation factor. With different dilation factors (i.e., $2.3$, $2.5$ and $2.7$ in this figure), we obtain different input features, and then get different output activations for the red location. Assuming the current dilation factor is $2.5$, we will get an output $15$. With a training signal, which expects the output activation increased or decreased, we can modify the current dilation factor along the direction to $2.3$ or $2.7$.}
\end{figure}

We extend the dilated convolution by introducing a dilation vector $d_c \in \mathbb{R}^{c+}$ for different channels, which can take fractional values.
The forward pass of the proposed dilated convolutions can be formulated as
\begin{equation}
\begin{split}
y_{m,n}  &=  \sum_{c}\sum_{i,j} w_{c,i,j}\cdot x_{c,m+i\cdot d_c,n+j\cdot d_c} + b.\\
\end{split}
\label{eq:channel_convolution}
\end{equation}
$x_{c,m+i\cdot d_c,n+j\cdot d_c}$ cannot directly be sampled from the input feature $\textbf{X}$ for most $d_c$. We obtain the value of a fractional position by employing bilinear interpolation on its four neighboring integer positions as shown in Fig.~2, specifically
\begin{equation}
\begin{split}
x&_{c,m+i\cdot d_c,n+j\cdot d_c} = \\
&x_{c, \lfloor m+i\cdot d_c\rfloor, \lfloor n+j\cdot d_c\rfloor} \cdot (1-\Delta d)^2  + x_{c, \lfloor m+i\cdot d_c\rfloor, \lceil n+j\cdot d_c\rceil} \cdot (1-\Delta d) \cdot \Delta d + \\
 & x_{c, \lceil m+i\cdot d_c\rceil, \lfloor n+j\cdot d_c\rfloor} \cdot \Delta d \cdot (1-\Delta d)  + 
  x_{c, \lceil m+i\cdot d_c\rceil, \lceil n+j\cdot d_c\rceil} \cdot (\Delta d)^2,\\
\end{split}
\label{eq:bilinear_interpolation}
\end{equation}
where $\Delta d = m+i\cdot d_c - \lfloor m+i\cdot d_c \rfloor = n+j\cdot d_c - \lfloor n+j\cdot d_c \rfloor$, is the decimal part of the dilation factor $d_c$.

\subsection{Backward pass}
\label{subsec:backward}
Chain rule and back propagation \cite{rumelhart1988learning} are used to optimize a deep neural network model. For each layer, networks obtain the training signals (gradients) from the next connected layer, and use them to update the parameters in current layer. Then, the processed gradient is passed to the previous connected layer.
Usually, those training signals are used to change the filter weights such that the output activation increases or decreases and the loss decreases.
Besides changing filter weights, changing the dilation factor provides another way to optimize a convolution networks, as discussed in Fig.~2.

The proposed dilated convolution layer based on bilinear interpolation is differentiable to dilation factors. Therefore, it allows us to train a full model end-to-end without any additional training signal for dilation factors.
Because $\Delta d$ in Eq.~\eqref{eq:bilinear_interpolation} is the decimal part of $c$-th channel's dilation factor $d_c$, the gradient for updating $d_c$ can be formulated as
\begin{equation}
\begin{split}
\frac{\partial L}{\partial d_c} &= \frac{\partial L}{\partial y_{m,n}} \cdot \frac{\partial y_{m,n}}{\partial d_c}
= \frac{\partial L}{\partial y_{m,n}} \cdot \frac{\partial y_{m,n}}{\partial \Delta  d}\\
&= \sum_{i,j} (x_{c, \lfloor m+i\cdot d_c\rfloor, \lfloor n+j\cdot d_c\rfloor} \cdot (2\cdot\Delta d-2) + 
  x_{c, \lfloor m+i\cdot d_c\rfloor, \lceil n+j\cdot d_c\rceil} \cdot (1-2\cdot\Delta d) + \\
 & \quad\qquad x_{c, \lceil m+i\cdot d_c\rceil, \lfloor n+j\cdot d_c\rfloor} \cdot (1-2\cdot\Delta d) + 
  x_{c, \lceil m+i\cdot d_c\rceil, \lceil n+j\cdot d_c\rceil} \cdot 2\cdot\Delta d)\\
  & \quad\qquad \cdot w_{c,i,j} \cdot\frac{\partial L}{\partial y_{m,n}}. \\
\end{split}
\label{eq:backward}
\end{equation}

Besides the dilation factors, filter weights and bias term also require updating. The gradient for the filter weights $\textbf{W}$ can be computed by
\begin{equation}
\begin{split}
\frac{\partial L}{\partial w_{c,i,j}} &= \frac{\partial L}{\partial y_{m,n}} \cdot \frac{\partial y_{m,n}}{\partial w_{c,i,j}}
= \frac{\partial L}{\partial y_{m,n}} \cdot x_{c,m+i\cdot d_c,n+j\cdot d_c},\\
\end{split}
\label{eq:filter_weight_update}
\end{equation}
where $x_{c,m+i\cdot d_c,n+j\cdot d_c}$ can be computed using Eq.~\eqref{eq:bilinear_interpolation}. The gradient for the bias term $\textbf{b}$ can be computed by
\begin{equation}
\begin{split}
\frac{\partial L}{\partial b} &= \frac{\partial L}{\partial y_{m,n}}.
\end{split}
\label{eq:bias_term_update}
\end{equation}

To employ back propagation for optimizing all the layers, the gradient for the input feature at location $(p,q)$ can be computed as the sum of the gradients from all the locations of output side, who sample the location $(p,q)$ in forward pass. The gradient at the output side $\frac{\partial L}{\partial y_{m,n}}$ will affect the gradients for all sampled input locations in bilinear interpolation. Specifically,
\begin{equation}
\begin{split}
\frac{\partial L}{\partial x_{c, \lfloor m+i\cdot d_c\rfloor, \lfloor n+j\cdot d_c\rfloor}} &= \frac{\partial L}{\partial y_{m,n}}\cdot w_{c,i,j}\cdot(1-\Delta d)^2, \\
\frac{\partial L}{\partial x_{c, \lfloor m+i\cdot d_c\rfloor, \lceil n+j\cdot d_c\rceil}} &= \frac{\partial L}{\partial y_{m,n}}\cdot w_{c,i,j}\cdot(1-\Delta d)\cdot\Delta d, \\
\frac{\partial L}{\partial x_{c, \lceil m+i\cdot d_c\rceil, \lfloor n+j\cdot d_c\rfloor}} &= \frac{\partial L}{\partial y_{m,n}}\cdot w_{c,i,j}\cdot\Delta d\cdot(1-\Delta d), \\
\frac{\partial L}{\partial x_{c, \lceil m+i\cdot d_c\rceil, \lceil n+j\cdot d_c\rceil}} &= \frac{\partial L}{\partial y_{m,n}}\cdot w_{c,i,j}\cdot(\Delta d)^2.
\end{split}
\label{eq:backwardx}
\end{equation}

\subsection{Network architectures}
\label{sec:network}

We select a set of state-of-the-art methods which use fixed and manually set dilation parameters. We employ our method and make the dilation parameters in those models learnable. Next, we describe the baseline models \cite{chen2016deeplab,pspnet2016zhao} in this paper and how we adopt them.

\mysubparagraph{Deeplab-LargeFOV} \cite{chen2016deeplab} is a VGG \cite{simonyan2014very} based semantic segmentation model. It replaces $\text{\it conv}5_1$, $\text{\it conv}5_2$ and $\text{\it conv}5_3$ in original VGG network with dilation $2$. And it has a convolution layer $\text{\it fc}6$ with $512$ input feature channels, $1024$ output channels and dilation $12$.
We make $\text{\it conv}5_1$, $\text{\it conv}5_2$, $\text{\it conv}5_3$ and 
$\text{\it fc}6$
learnable, and set the range of them as $[1,4]$, $[1,4]$, $[1,4]$ and $[4,20]$.

\mysubparagraph{Deeplab-v2} \cite{chen2016deeplab} uses ResNet-101 \cite{he2016deep} to extract visual features. It modifies the original ResNet-101 with dilated convolutions. Before $\text{\it res}5{\it c}$, there are 23 layers with dilation 2, and 3 layers with dilation 4. There is an ASPP layer after $\text{\it res}5{\it c}$, which combines dilation 6, 12, 18 and 24 to recognize the class of each location.
We set the range of the dilated convolution layer with dilation factor 2, 4, 6, 12, 18 and 24 to $[1,4]$, $[1,8]$, $[1,11]$, $[7,17]$, $[13,23]$ and $[19,29]$, respectively.

\mysubparagraph{PSPNet} \cite{pspnet2016zhao} has a similar architecture to \textit{Deeplab-v2}, and achieves state-of-the-art performance on various datasets. There are 23 layers with dilation 2, and 3 layers with dilation 4.
Similarly, we set the range of the dilated convolution layer with dilation factor 2 and 4 to $[1,4]$ and $[1,8]$.
We show that our method can boost \textit{PSPNet}, and achieve new state-of-the-art performance on the challenging street view dataset Cityscapes \cite{cordts2016cityscapes}.
\begin{table}[t]
\tiny
  \begin{center}
    \caption{Hyperparameters in the experiments of this paper. }
    \label{table:hyperparameters}
    \begin{tabular}{lccccc}
      \toprule
       Datasets & Networks & Batch size & \tabincell{c}{Image patch \\ size}  & \tabincell{c}{Base \\ learning rate} & iterations \\
      \cmidrule(lr){1-6}
                   & Deeplab-LargeFOV  & 10 & 321$\times$ 321 & $1\times 10^{-3}$  & 20,000 \\
      Cityscapes   & Deeplab-v2  & 10 & 321$\times$ 321 & $2.5\times 10^{-4}$  & 20,000 \\
      & PSPNet & 16 & 561$\times$ 561 & $1\times 10^4$  & 20,000 \\
      \cmidrule(lr){1-6}
      CamVid   & Deeplab-v2  & 10 & 321$\times$ 321 & $2.5\times 10^{-4}$  & 15,000  \\
      & PSPNet  & 10 & 473$\times$ 473 & $1\times10^{-4}$   & 10,000  \\    
      \bottomrule
    \end{tabular}
  \end{center}
\end{table}

\subsection{Implementation details}
\label{subsec:implementation}
We train our Deeplab-LargeFOV, Deeplab-v2 and PSPNet models with filter weights initialized from released, original models, which are trained on PASCAL VOC 2012 \cite{everingham2010pascal}, MS-COCO \cite{coco2014} and Cityscapes \cite{cordts2016cityscapes}, respectively. We initialize the dilation factors with the manually set value of the original dilated convolutions. We apply batch SGD with momentum to optimize the models. The batch size is set to 10, the momentum is 0.9 and the weight decay is 0.0005. We use the ``poly" learning rate policy where the current learning rate is equal to the base learning rate multiplying
$(1-\frac{iter}{iter_{max}})^{power}$, and $power=0.9$ for all the experiments. The other hyperparameters are presented in Tab.~\ref{table:hyperparameters}.
We use the respective training code released from the authors. We implement our method using the \textit{Caffe} \cite{jia2014caffe} framework and the source code is available at \url{https://github.com/SSAW14/LearnableDilationNetwork}. 
In our experiments, our models need 3\% to 8\% additional time in inference. 
Training Deeplab-LargeFOV, Deeplab-v2 and PSPNet need only additional ~10\%, ~15\% and ~15\% computational time compared to the base models, respectively.


    \section{Experiments}
    \label{sec:exp}
We evaluate the proposed method and baselines on the public benchmarks Cityscapes \cite{cordts2016cityscapes} and CamVid \cite{camvid2008} using four evaluation metrics following previous work \cite{long2015fully,yang_cvpr17}:
pixel accuracy (\textit{Pixel Acc.}), mean class accuracy (\textit{Cls Acc.}), region intersection over union (\textit{Mean IoU}),
and frequency weighted intersection over union (\textit{f.w. IoU}).

\subsection{Cityscapes}
\label{subsec:citycapes}
Cityscapes \cite{cordts2016cityscapes} is a recently released street scene dataset, which is collected from diverse cities in different seasons. The image resolution in Cityscapes is $1024\times 2048$ and the image quality is very high. It defines 19 semantic classes covering traffic, stuff and objects. There are 2975, 500 and 1525 carefully annotated images for training, validation and testing. Besides, there are also 20,000 coarsely annotated images provided for additional training data. Following previous work \cite{pspnet2016zhao}, we leverage those coarse annotations during training to obtain state-of-the-art performance.



\myparagraph{Ablation study for Deeplab-LargeFOV.}
We first provide an ablation study on the Cityscapes validation 
set to show the effectiveness of learning dilation factors using the baseline model Deeplab-LargeFOV. We train models with different dilation configurations (see Tab.~\ref{table:largefov}), varying fixed dilations in $\text{conv}5_1$ to $\text{conv}5_3$ from 1 to 4, and learning parameters for $\text{conv}5_1$ to $\text{conv}5_3$ and $\text{fc}6$.

The first row in Tab.~\ref{table:largefov} shows the performance of the baseline method without dilated convolutions, which is only 58.91\% mean IoU. Fixed integer valued dilated convolution parameters can improve the performance to up to 62.51\% for a factor of 4 in $\text{conv}5_1$ to $\text{conv}5_3$.


\begin{table}[!b]
\scriptsize
  \begin{center}
    \caption{Ablation study on the Cityscapes validation set using the VGG based Deeplab-LargeFOV model. Black numbers for the convolutional layers indicate fixed dilation parameters, {\color{red}red} numbers or ranges in our learnable dilated convolution layers indicate the initial values or distributions before training. }
    \label{table:largefov}
    \begin{tabular}{ccccccccccccc}
      \toprule
      & $\text{conv}5_1$ & $\text{conv}5_2$ & $\text{conv}5_3$ & $\text{fc}6$ & & \textit{Pixel Acc.} & \textit{Cls Acc.} & \textit{f.w. IoU} & \textit{Mean IoU} & learned $\text{conv}5_{1-3}$&learned  $\text{fc}6$\\
      \cmidrule(lr){1-12}
      & 1 & 1 & 1 & 12 & & 93.11 & 68.70 & 87.76& 58.91  & &\\
      & 2 & 2 & 2 & 12 & & 93.49 & 71.76 & 88.36& 61.44  & &\\
      & 3 & 3 & 3 & 12 & & 93.50 & 71.93 & 88.37& 62.17  & &\\
      & 4 & 4 & 4 & 12 & & 93.49  & 72.35  & 88.38& 62.51    & &\\
      & 2.35 & 2.6 & 3.5 & 12 & & 93.63  & 72.46  & 88.58& 62.94    && \\
      \cmidrule(lr){1-12}
& {\color{red}2} & {\color{red}2} & {\color{red}2} & 12 & & 93.50 & 73.25 & 88.41& 62.31  & \checkmark&\\
      & 2 & 2 & 2 & {\color{red}12} & & 93.65 & 72.64 & 88.61 & 62.86 & &\checkmark\\
      & {\color{red}[1,4]} & {\color{red}[1,4]} & {\color{red}[1,4]} & {\color{red}[4,20]} & & {93.69} & {72.90}  & {88.71}& {62.92} & \checkmark&\checkmark\\
      & {\color{red}2} & {\color{red}2} & {\color{red}2} & {\color{red}12} & & \textbf{93.72} & \textbf{73.38}  & \textbf{88.77}& \textbf{63.31} & \checkmark&\checkmark\\
      \bottomrule
    \end{tabular}
  \end{center}
\end{table}

\begin{figure}[t]
\begin{center}
   \includegraphics[width=0.99\linewidth]{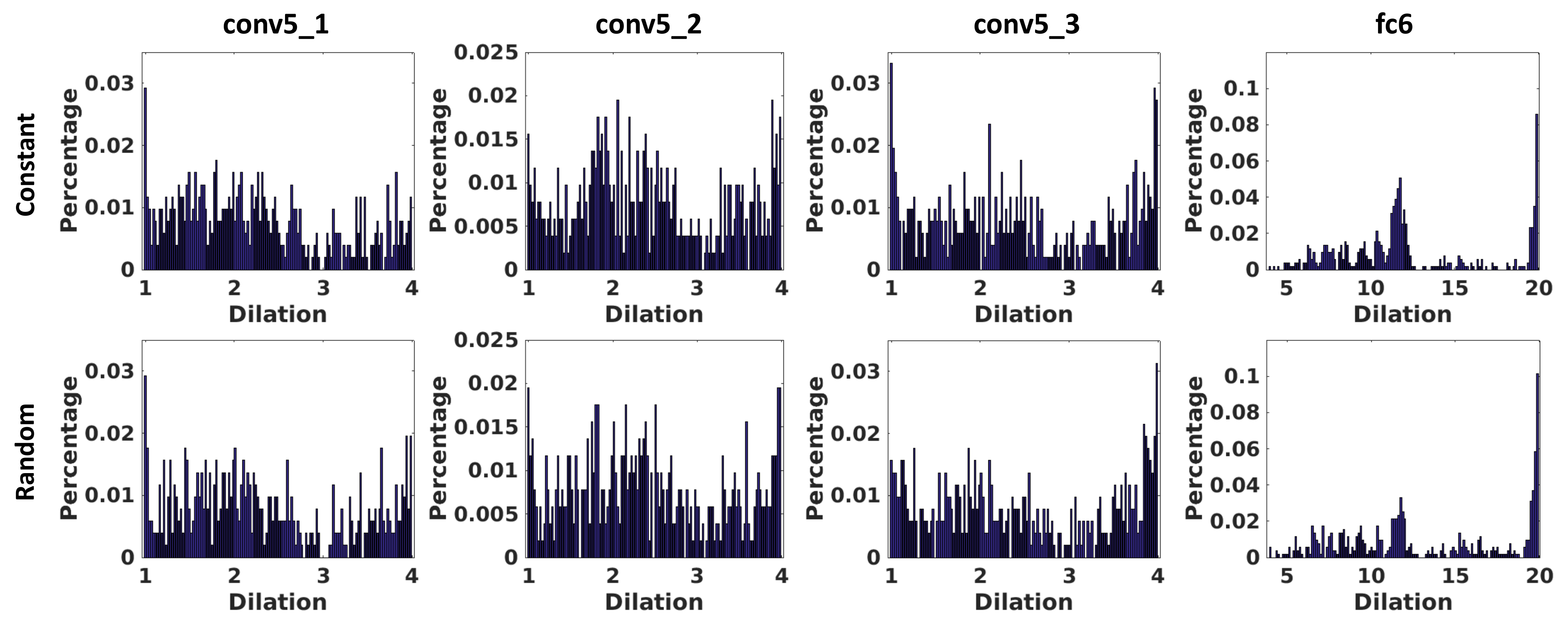}
   \label{fig:distribution_deeplab_large}
\end{center}
\caption{The learned dilation distribution in the Deeplab-LargeFOV model on Cityscapes dataset. The first row shows the distribution using constant value initialization, and the second row shows the distribution using random noise as presented in Tab.~\ref{table:largefov}.}
\end{figure}

By replacing the fixed dilation parameters with our learnable dilated convolutions using the same dilation factors as initialization, we get a further improvement to up to 63.31\% mean IoU.
Besides, we also use uniform distribution for the initialization of dilated convolutions, obtaining 62.92\% mean IoU, which is comparable to constant value initialization, and better than the models with fixed dilation factors.
%
We show the dilation distribution over input channels in Fig.~3. We observe that the learned dilation distributions of using constant value and random noise, are very similar, which is clearly shown the stability of optimization.
We observe that our dilation covers most values in the range of $[1,4]$ for $\text{conv}5_1$ to $\text{conv}5_3$, which allows us capture \textit{local} details and \textit{wide} context at the same time. The second observation is that there are some peaks in the distribution, which make the current layer capture more local information or capture a wider context. 
For a specific dilated convolution layer in a network, it is very difficult to know whether the current convolution should capture local or global information, which is why the proposed learning based model achieves better performance than all fixed dilation settings.
To verify this point, we train a further model with fixed dilated convolutions with factors $2.35$, $2.6$ and $3.5$ which are the average values of the learned dilation distribution over all channels. This model achieves better results than all the other fixed settings but is worse than the channel-wise learned setting.


\myparagraph{Comparisons on Deeplab-v2 and PSPNet.}
We choose Deeplab-v2 and PSPNet as our baselines, which leverage the powerful ResNet \cite{he2016deep} to build their models. We report the \textit{Mean IoU} score. The comparison results on Cityscapes validation set can be found in Tab.~3. Compared our results to baselines, we observe that we got general improvement in most classes like ``Fence'', ``Terrain'', ``Car'' and ``Bus''.
Particular, in some challenging and important (core role in traffic scenarios) classes like ``Person'' and ``Rider'', we got clearly improvements for Deeplab-v2 as well as PSPNet.
Due to deformations and large appearance variances, ``Person'' and ``Rider'' are easily confused each other.
By learning appropriate details and context, we boosted Deeplab-v2 and PSPNet to recognize ``Person'' (+0.6 percentage points ($pp$) for Deeplab-v2 and +1.3$pp$ for PSPNet) and ``Rider'' (+0.8$pp$ for Deeplab-v2 and +4.9$pp$ for PSPNet).
Fig.~4 shows some qualitative results from baselines and our models. In the left two columns, Deeplab-v2 baseline recognizes the rider and the middle section of the train to classes ``person'' and ``bus'', which are confusing to ground truth. In the right two columns, PSPNet baseline fails to recognize the font part of the truck and the rider, while our method shows improved predictions.

\begin{table}[t]
\tiny
  \begin{center}
    \label{table:cityscapes_comparison}
 \begin{tabular}{clccccccccccccccccccccccc}
      \toprule
       & Method &  &  
       & & \rotatebox{75}{Road} & \rotatebox{75}{Sidewalk} & \rotatebox{75}{Building} & \rotatebox{75}{Wall} & \rotatebox{75}{Fence} & \rotatebox{75}{Pole} & \rotatebox{75}{Traf. Light} & \rotatebox{75}{Traf. Sign} & \rotatebox{75}{Vegetation} & \rotatebox{75}{Terrain} & \rotatebox{75}{Sky} & \rotatebox{75}{Person} & \rotatebox{75}{Rider} &  \\ 
      \cmidrule(lr){1-19}
      &   Deeplab-v2 \cite{chen2016deeplab} &  &   & & 97.2 & 78.7 & 90.2 & 49.3 & 48.8 & 52.5 & \textbf{57.7} & 69.7 &  90.8 & 59.4   & 92.7 & 76.6 & 53.3 &  \\
      &   Deeplab-v2 + ours  &  &  & & 97.2 & \textbf{79.1} & \textbf{90.5} & \textbf{52.2} & \textbf{49.9} & \textbf{53.2} & 57.4 & \textbf{70.1} &  \textbf{90.9} & \textbf{59.6}   & \textbf{92.9} & \textbf{77.2} & \textbf{54.6} & \\
	  \cmidrule(lr){1-19}
      &   PSPNet \cite{pspnet2016zhao}  &  &  &   & 98.3 & 86.4 & \textbf{93.1} & \textbf{60.6} & 65.9 & \textbf{64.3} & 72.0 & 81.1 &  92.6 & 64.7   & 94.9 & 82.5 & 61.5 &  \\
      &   PSPNet + ours &  &  &  &  98.3 & 86.4 & 93.0 & 59.1 & \textbf{66.4} & 64.0 & \textbf{72.7} & \textbf{81.3} &  92.6 & \textbf{65.6}   & 94.9 & \textbf{83.3} & \textbf{66.4} &  \\
\hline
    \end{tabular}
    
\begin{tabular}{clccccccccccccccc}
   & Method &  &  & & \rotatebox{75}{Car} & \rotatebox{75}{Truck} &  \rotatebox{75}{Bus}&  \rotatebox{75}{Train} & \rotatebox{75}{Motorcycle} & \rotatebox{75}{Bicycle} & & \rotatebox{75}{\textit{Mean IoU}} & \\
   \cmidrule(lr){1-14}
    & Deeplab-v2 \cite{chen2016deeplab} &  &  & & 92.6 & \textbf{66.8} & 78.1 & \textbf{61.3} & \textbf{60.4} & 71.9 & & 70.9 & \\
       & Deeplab-v2 + Ours &  &  & & \textbf{92.8} & 62.8 & \textbf{79.9} & 58.9 & 58.9 & \textbf{72.2} & & \textbf{71.1} & \\
       	  \cmidrule(lr){1-14}
           & PSPNet \cite{pspnet2016zhao} &  &  & & 95.3 &  81.4 & 89.7 & \textbf{84.5} & 62.8 & 78.7 & & 79.4 & \\
       & PSPNet + Ours &  &  & &\textbf{95.4} &  \textbf{83.0} & \textbf{89.9} &   80.6 & \textbf{66.8} & \textbf{78.9} & & \textbf{79.9} & \\
        \bottomrule    
\end{tabular}
  \end{center}
\caption{Comparison IoU scores on Cityscapes validation set.}
\end{table}

\begin{figure*}[!t]
\centering
\scalebox{1}{
\centering
\begin{tabular}{@{}l@{\hspace{0.03cm}}l@{\hspace{0.02cm}}l@{\hspace{0.04cm}}l@{\hspace{0.02cm}}l@{}}\multicolumn{1}{c}{} & \multicolumn{2}{c}{\textbf{Deeplab-v2 \cite{chen2016deeplab}}}&\multicolumn{2}{c}{\textbf{PSPNet \cite{pspnet2016zhao} }}\\
\rotatebox{90}{data}&
\includegraphics[width=0.22\linewidth]{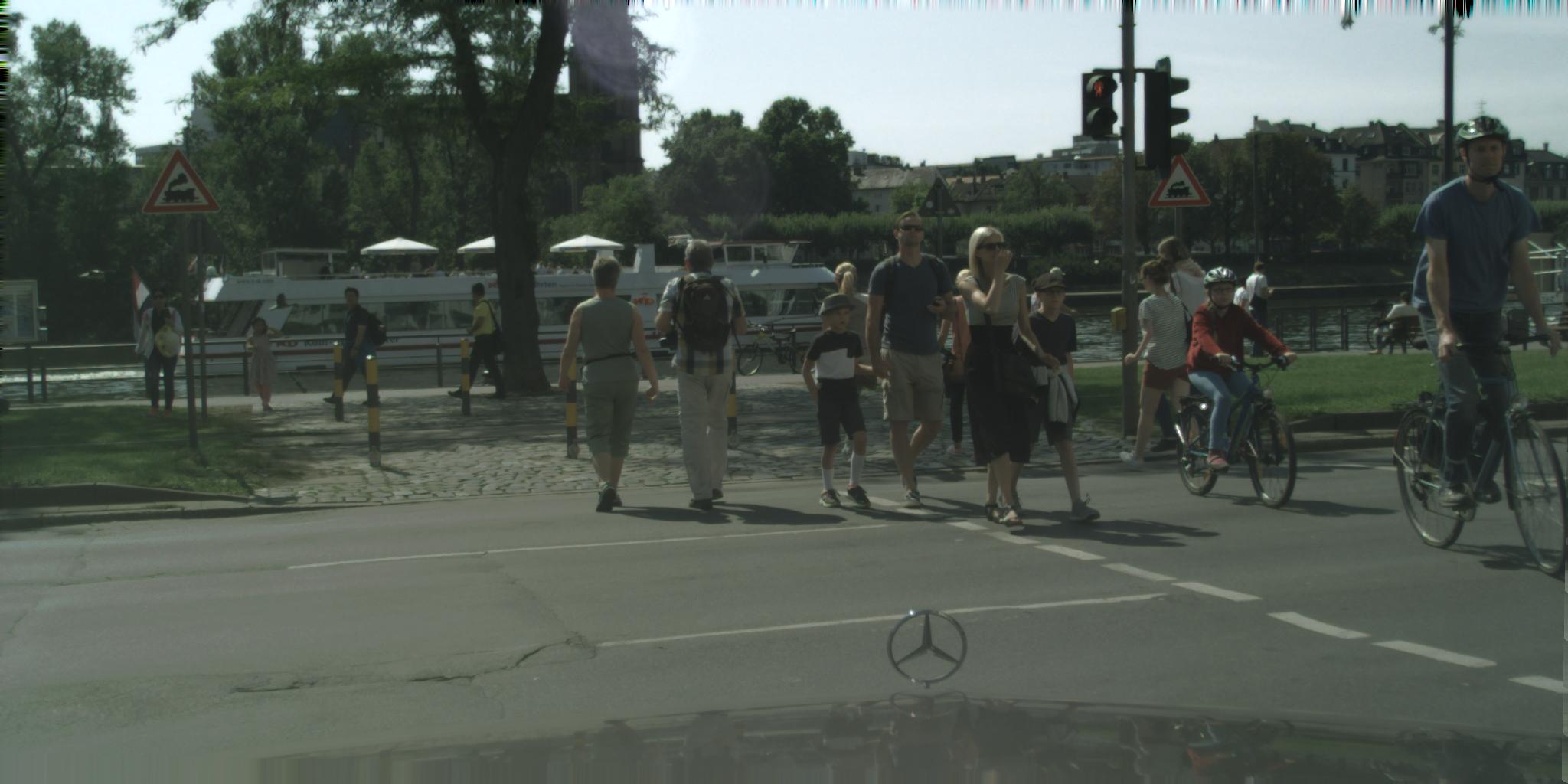}&
\includegraphics[width=0.22\linewidth]{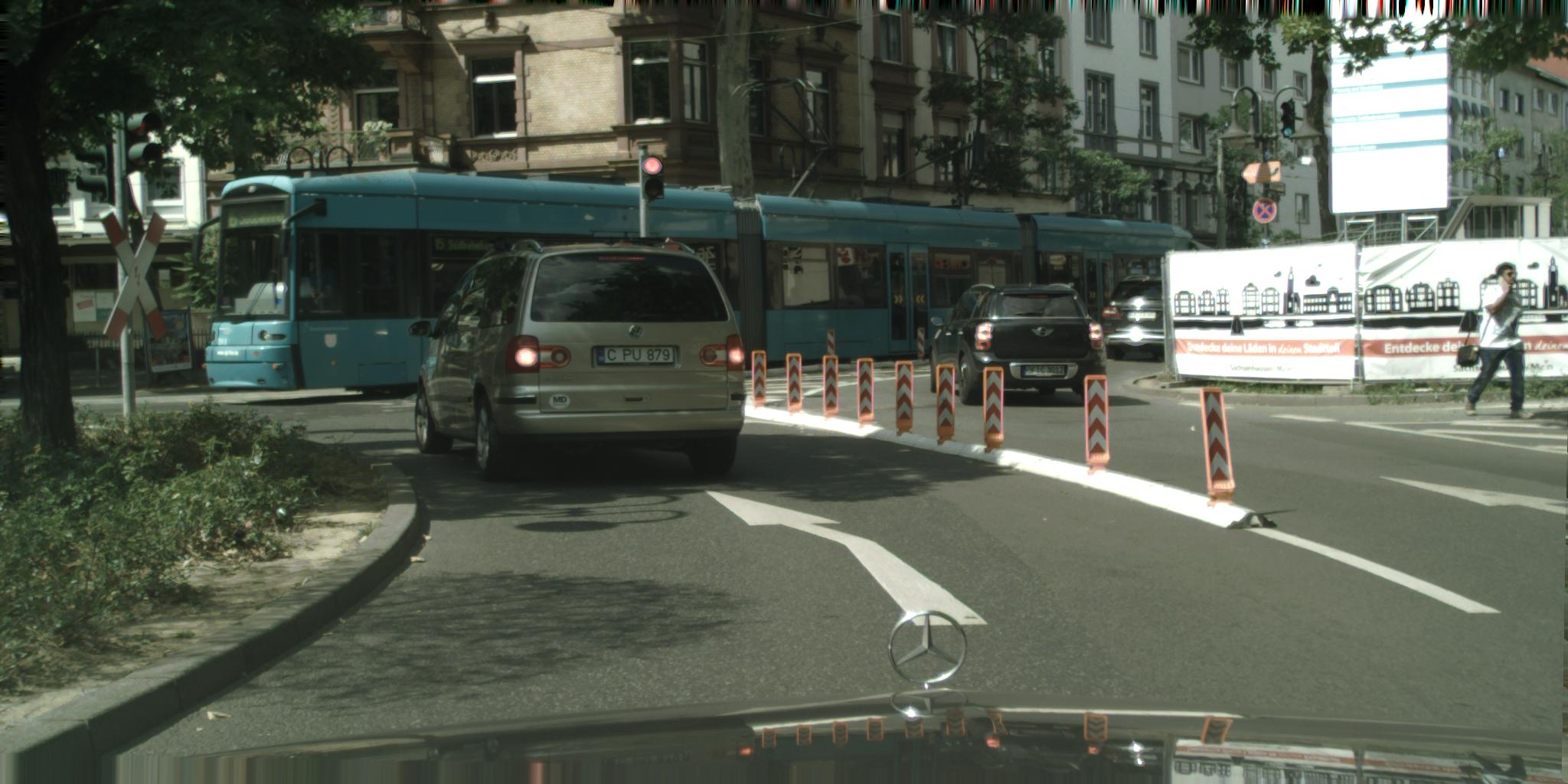}&
\includegraphics[width=0.22\linewidth]{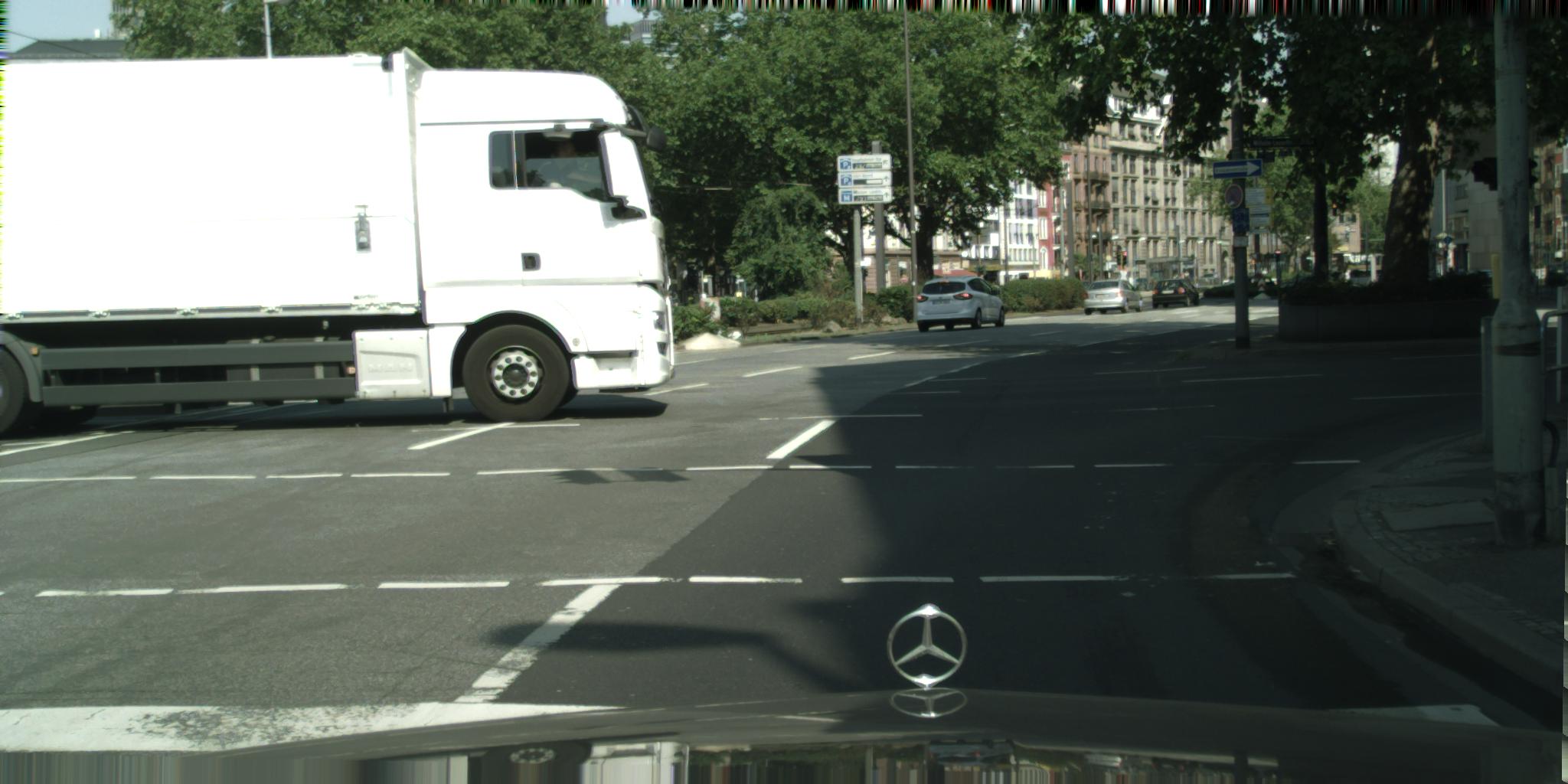}&
\includegraphics[width=0.22\linewidth]{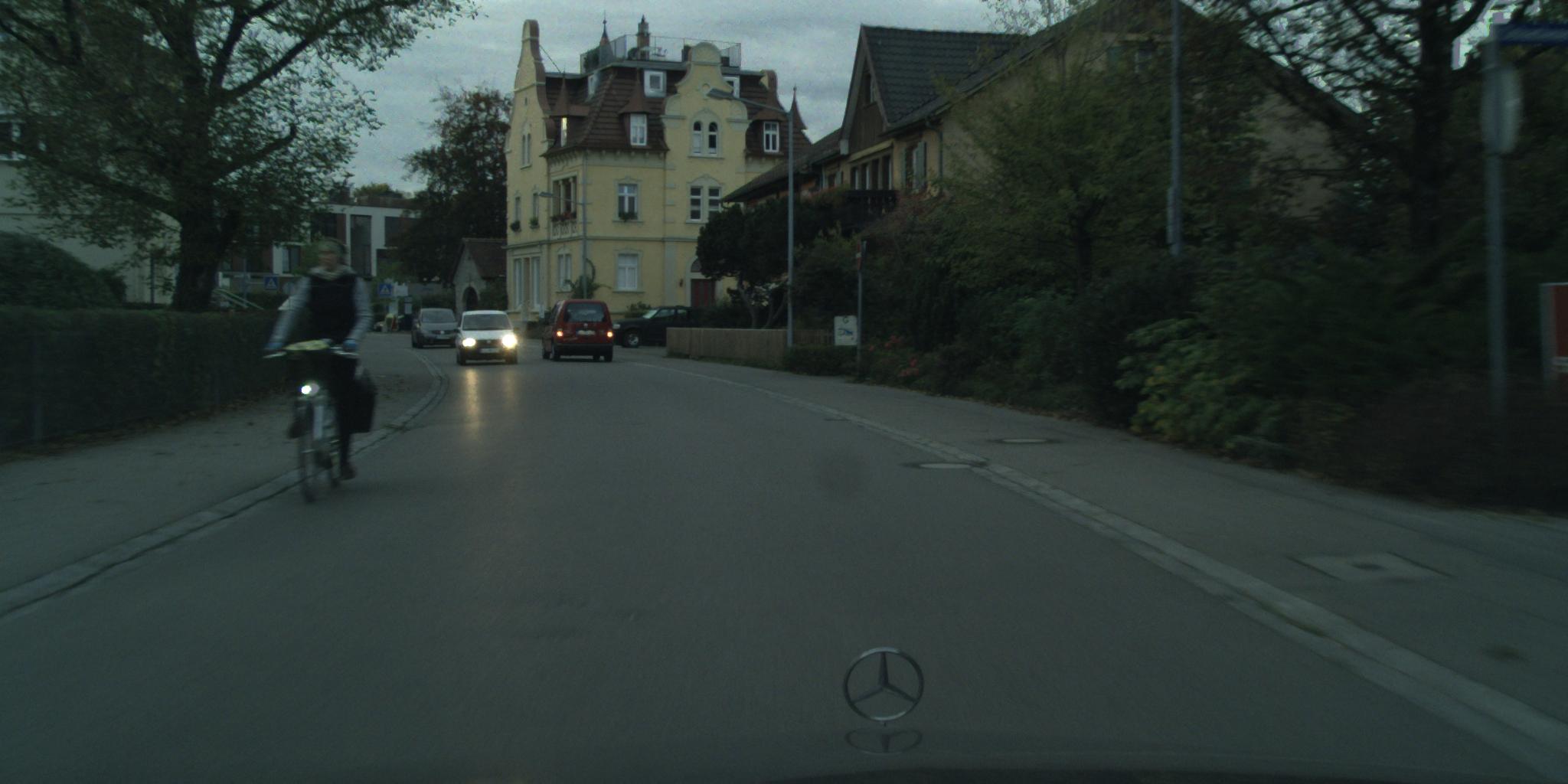}\\
\rotatebox{90}{GT}&
\includegraphics[width=0.22\linewidth]{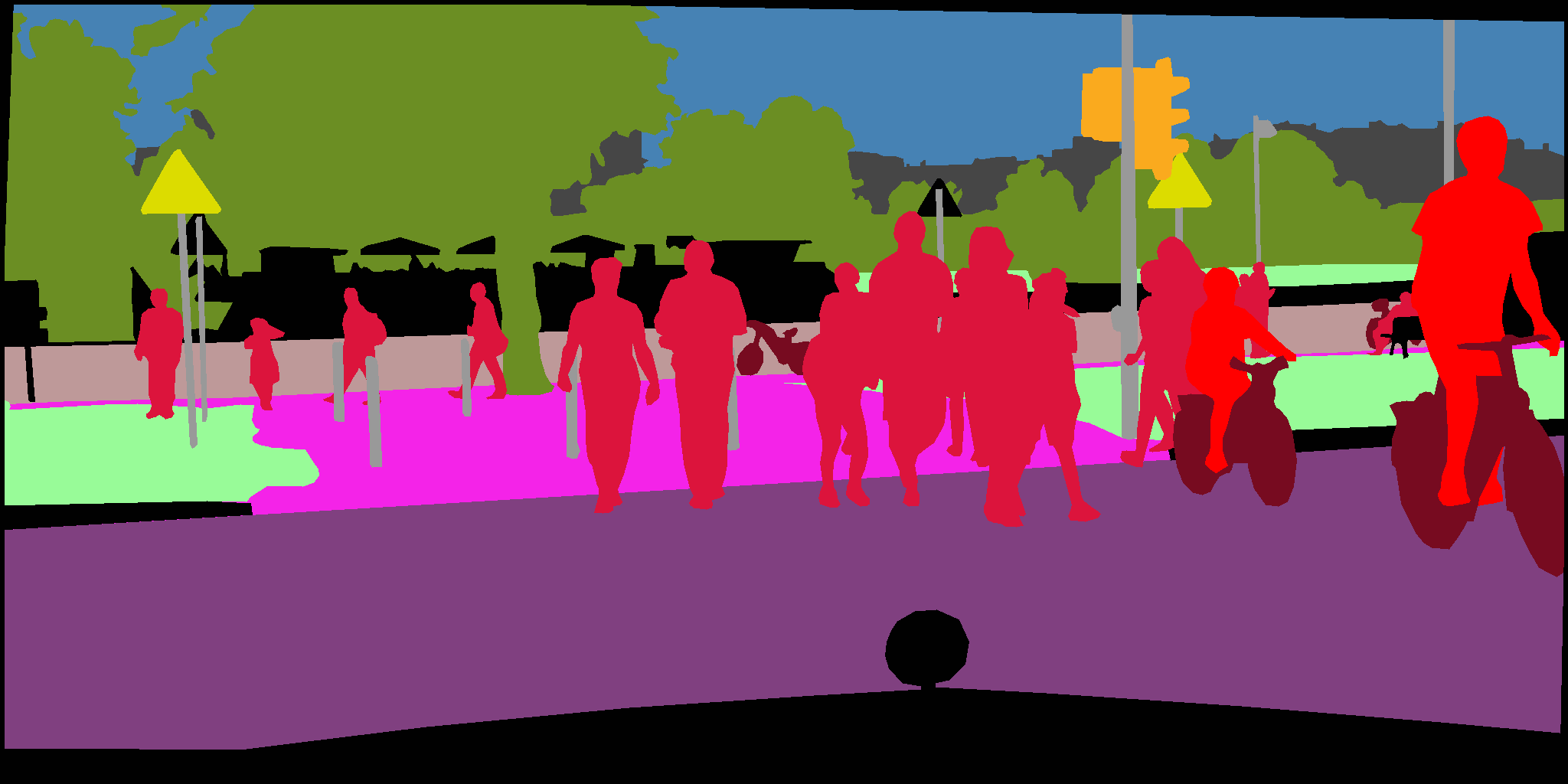}&
\includegraphics[width=0.22\linewidth]{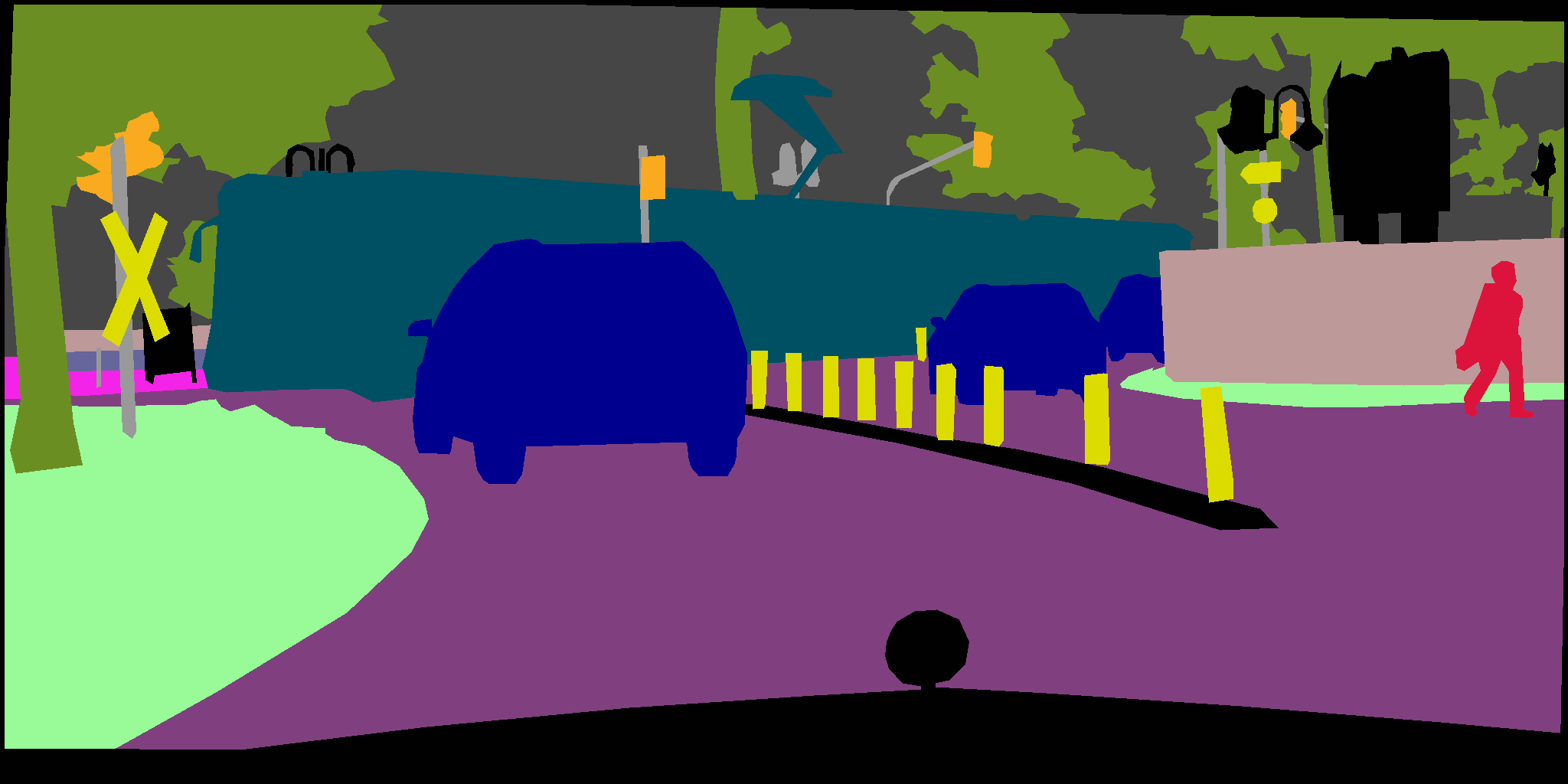}&
\includegraphics[width=0.22\linewidth]{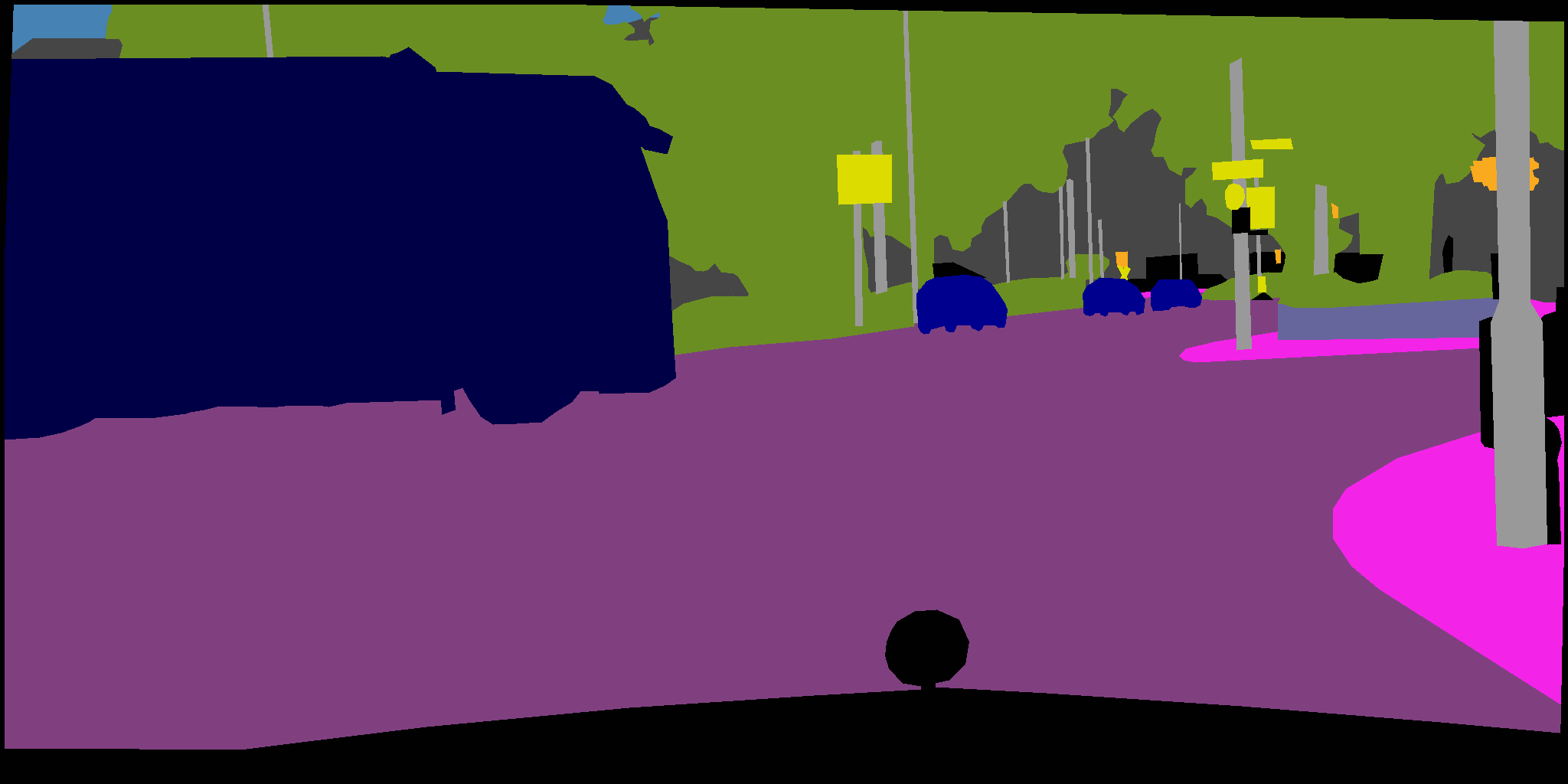}&
\includegraphics[width=0.22\linewidth]{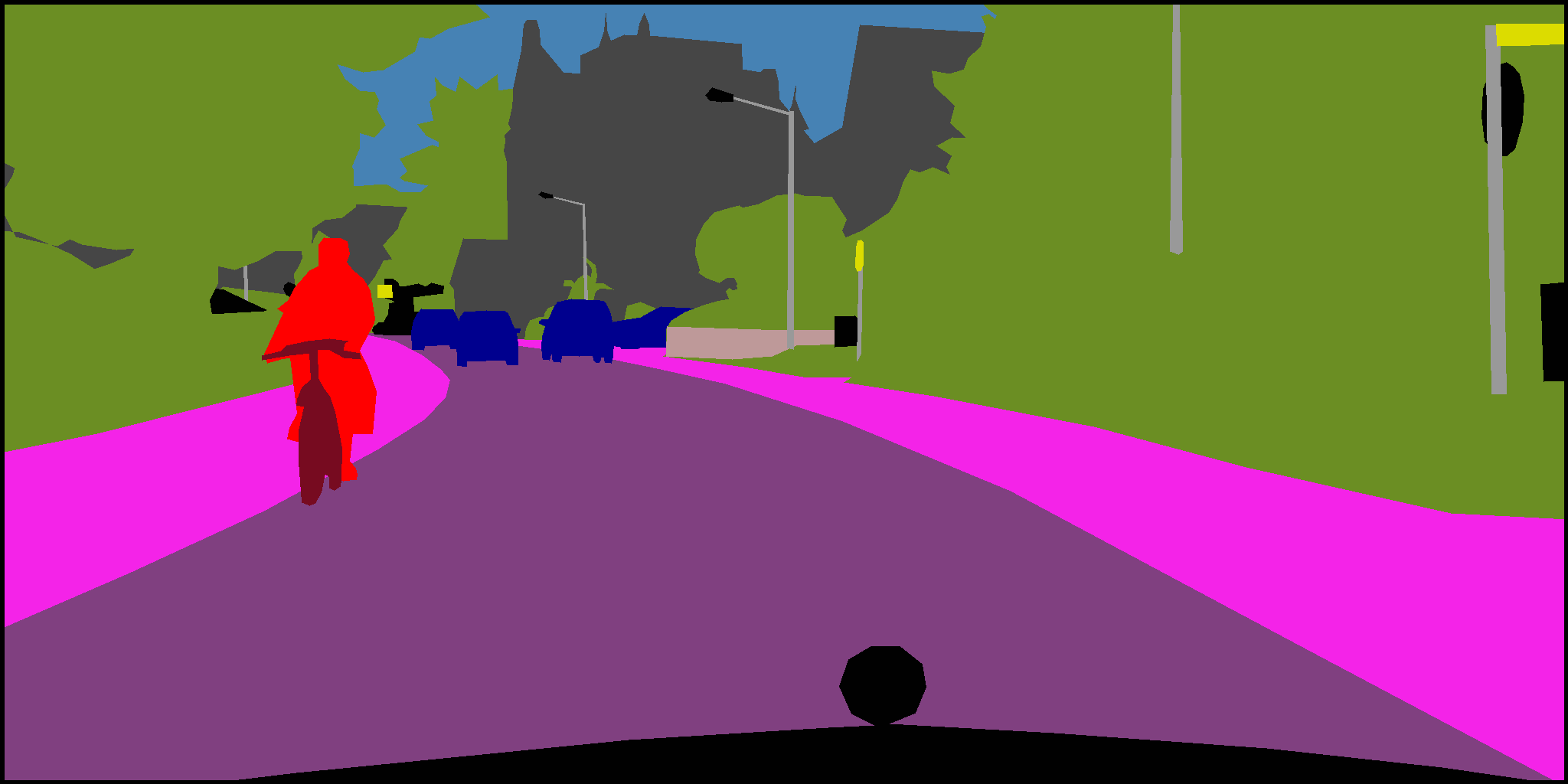}\\
\rotatebox{90}{baseline}&
\includegraphics[width=0.22\linewidth]{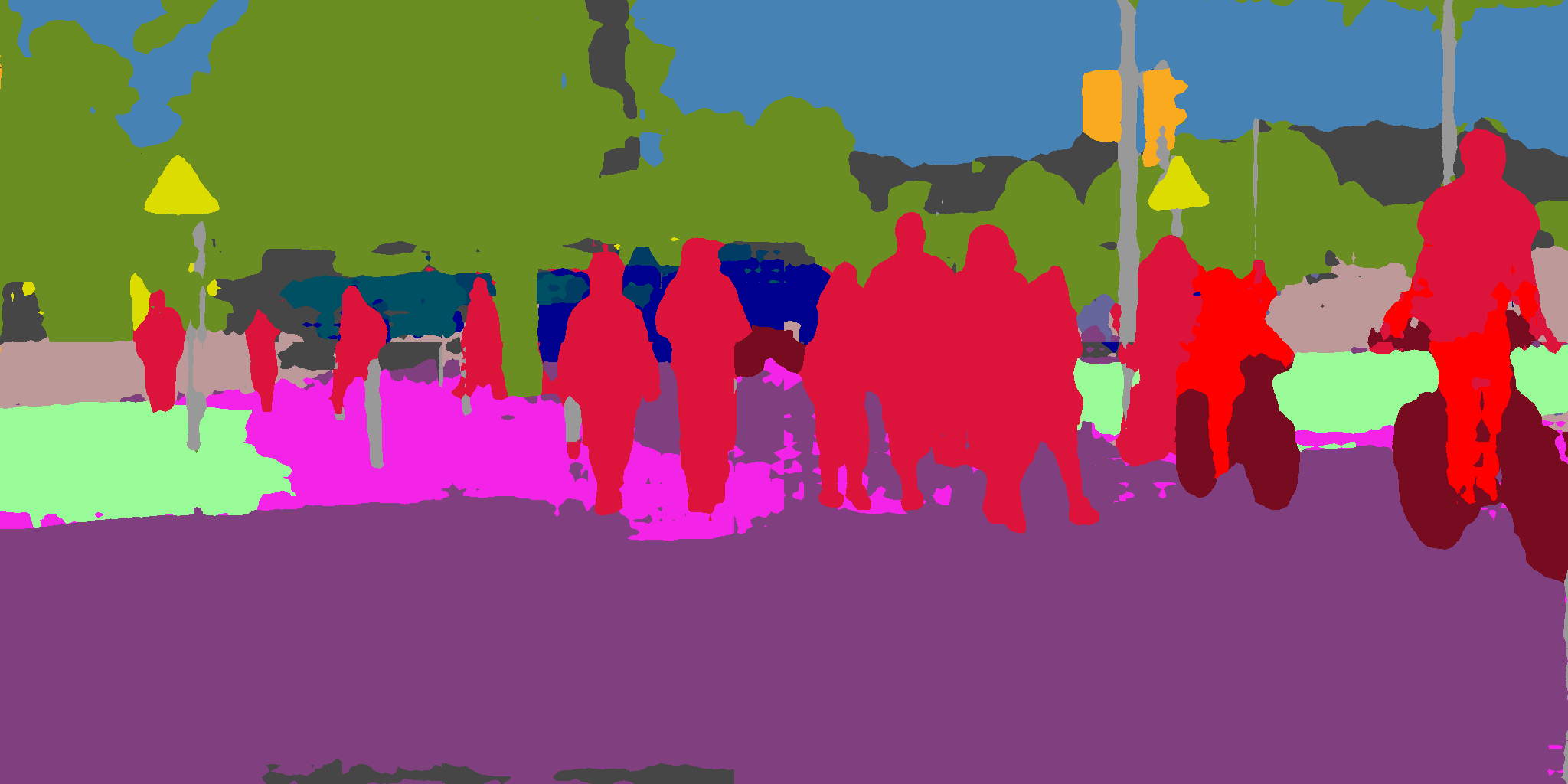}&
\includegraphics[width=0.22\linewidth]{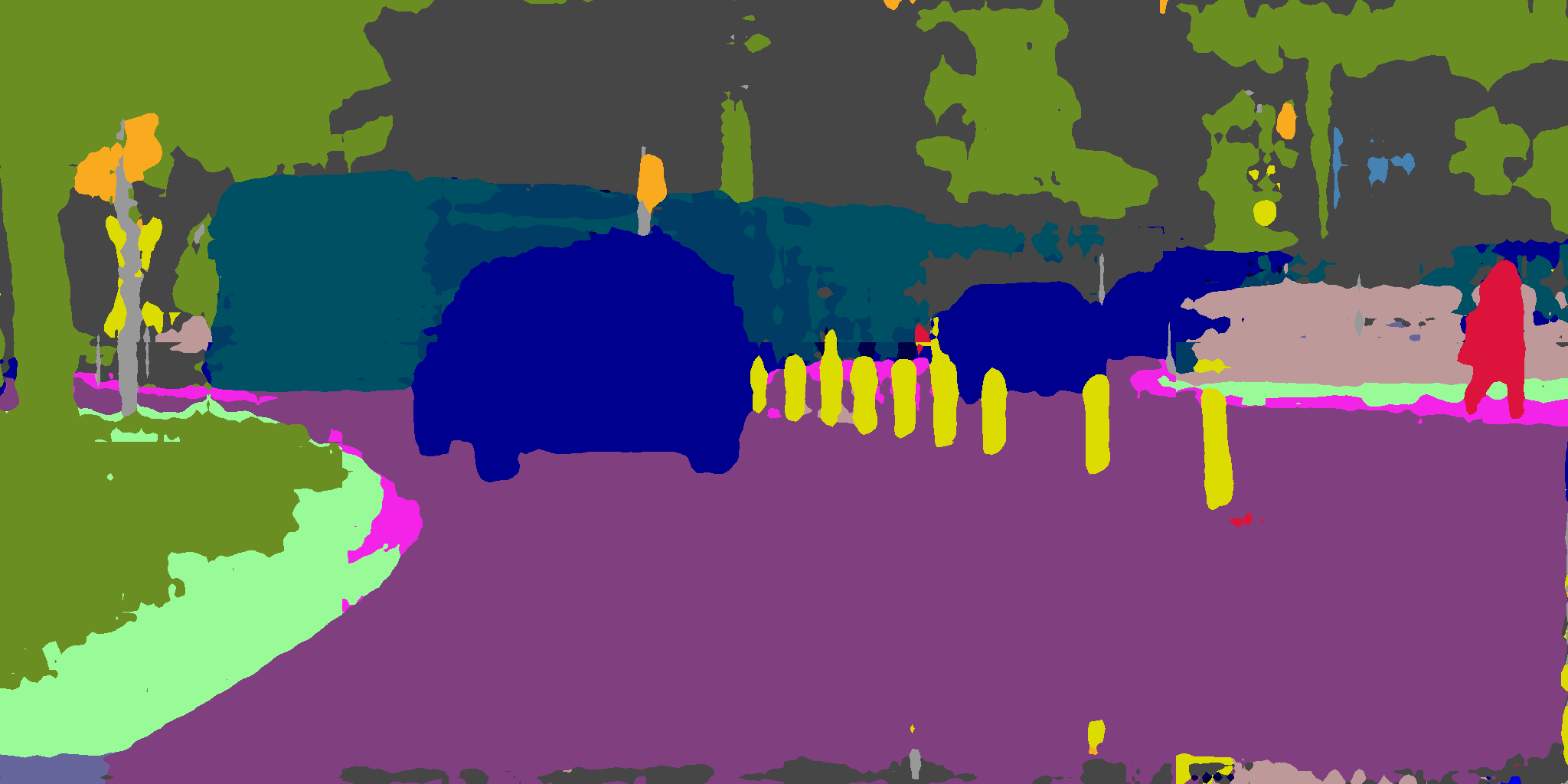}&
\includegraphics[width=0.22\linewidth]{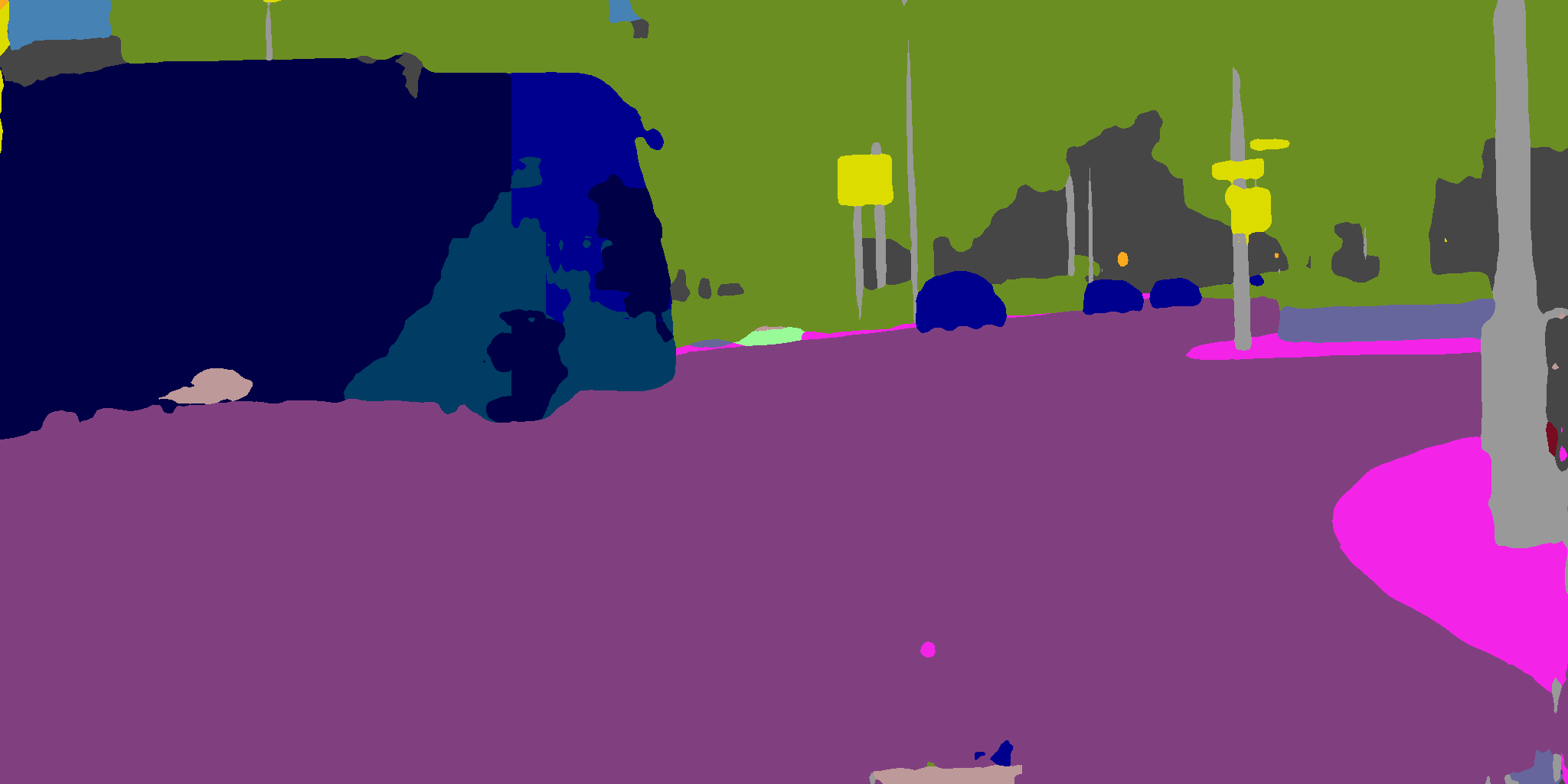}&
\includegraphics[width=0.22\linewidth]{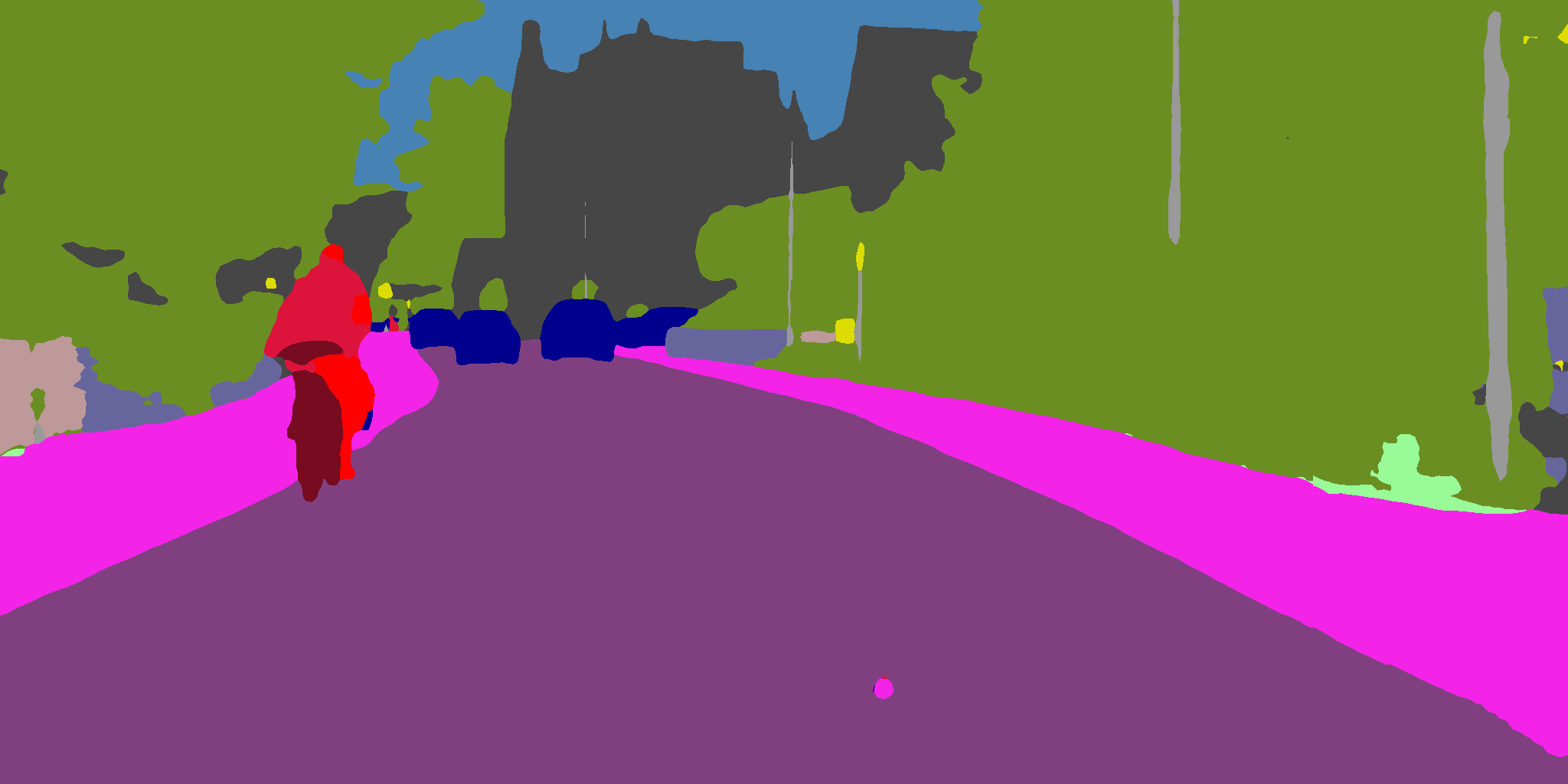}\\
\rotatebox{90}{Ours}&
\includegraphics[width=0.22\linewidth]{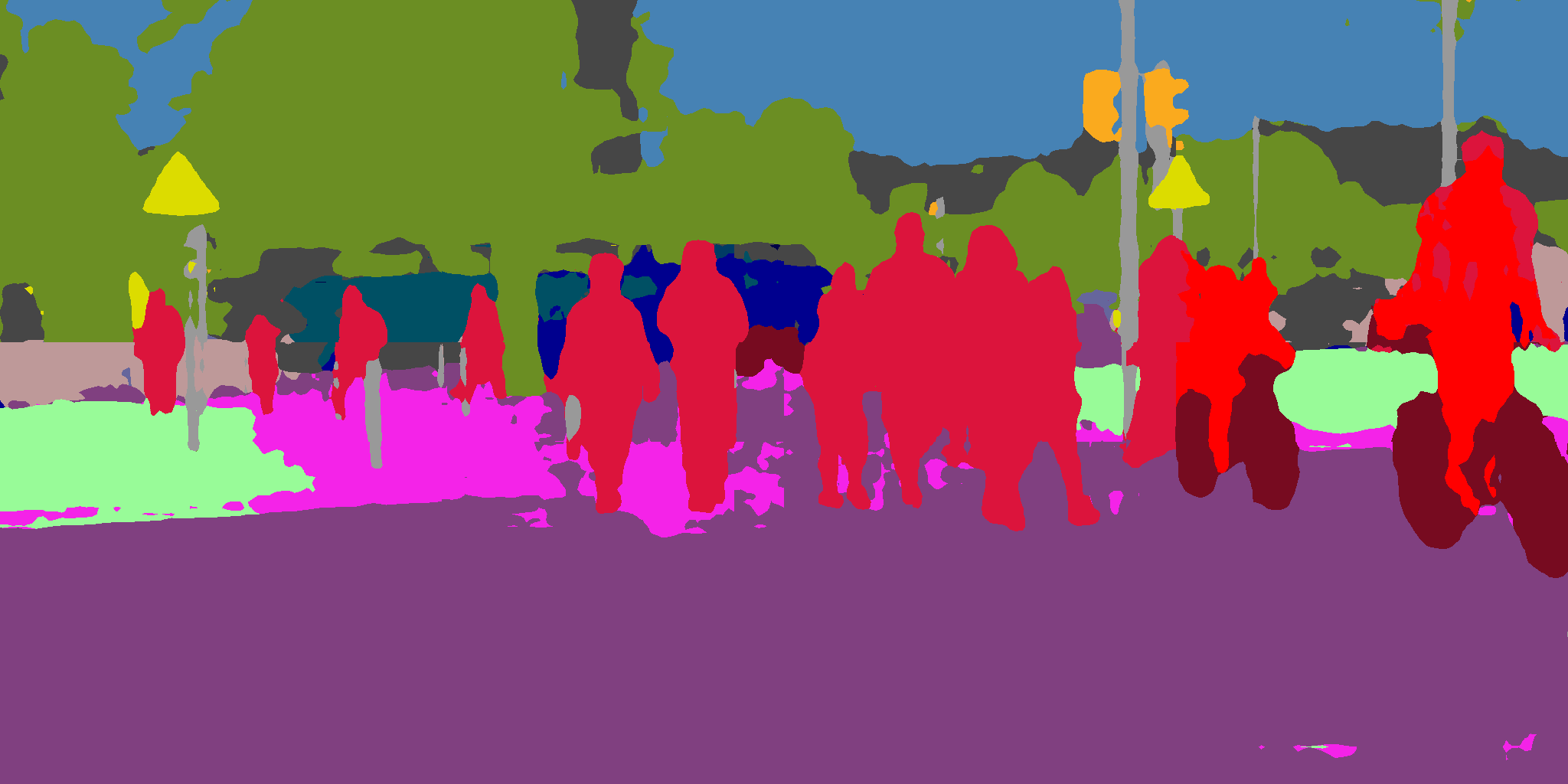}&
\includegraphics[width=0.22\linewidth]{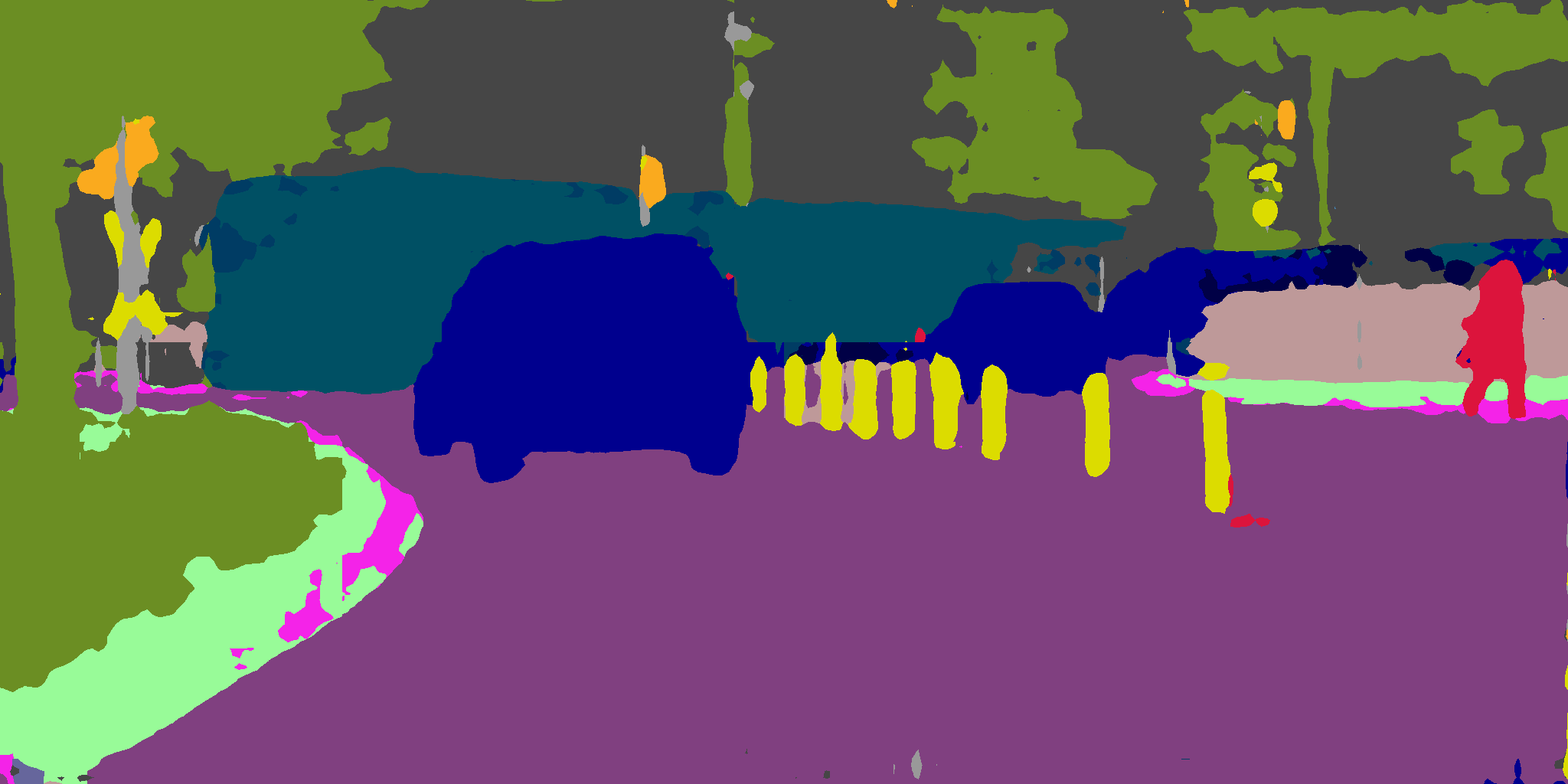}&
\includegraphics[width=0.22\linewidth]{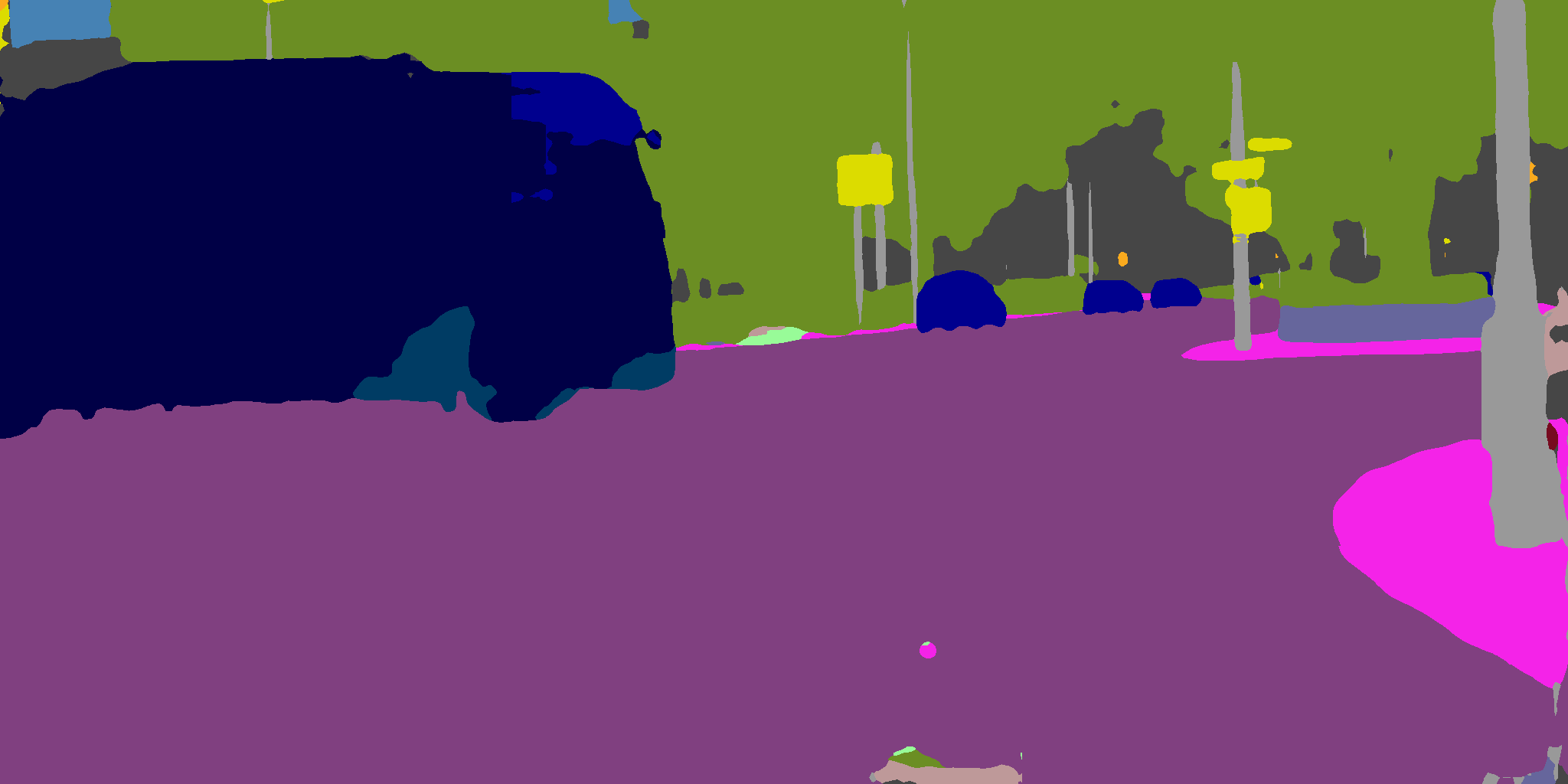}&
\includegraphics[width=0.22\linewidth]{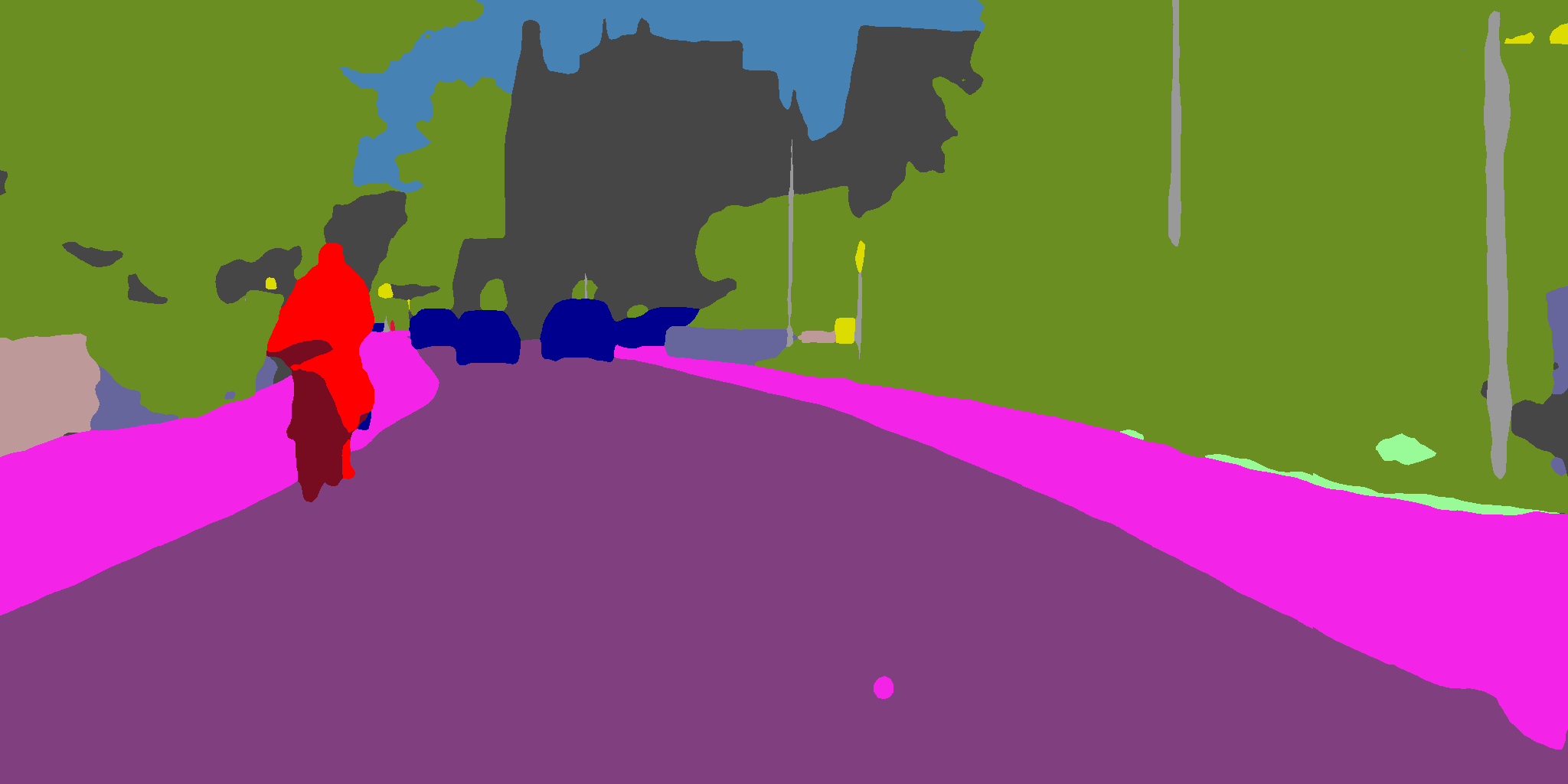}\\
\rotatebox{90}{Difference}&
\includegraphics[width=0.22\linewidth]{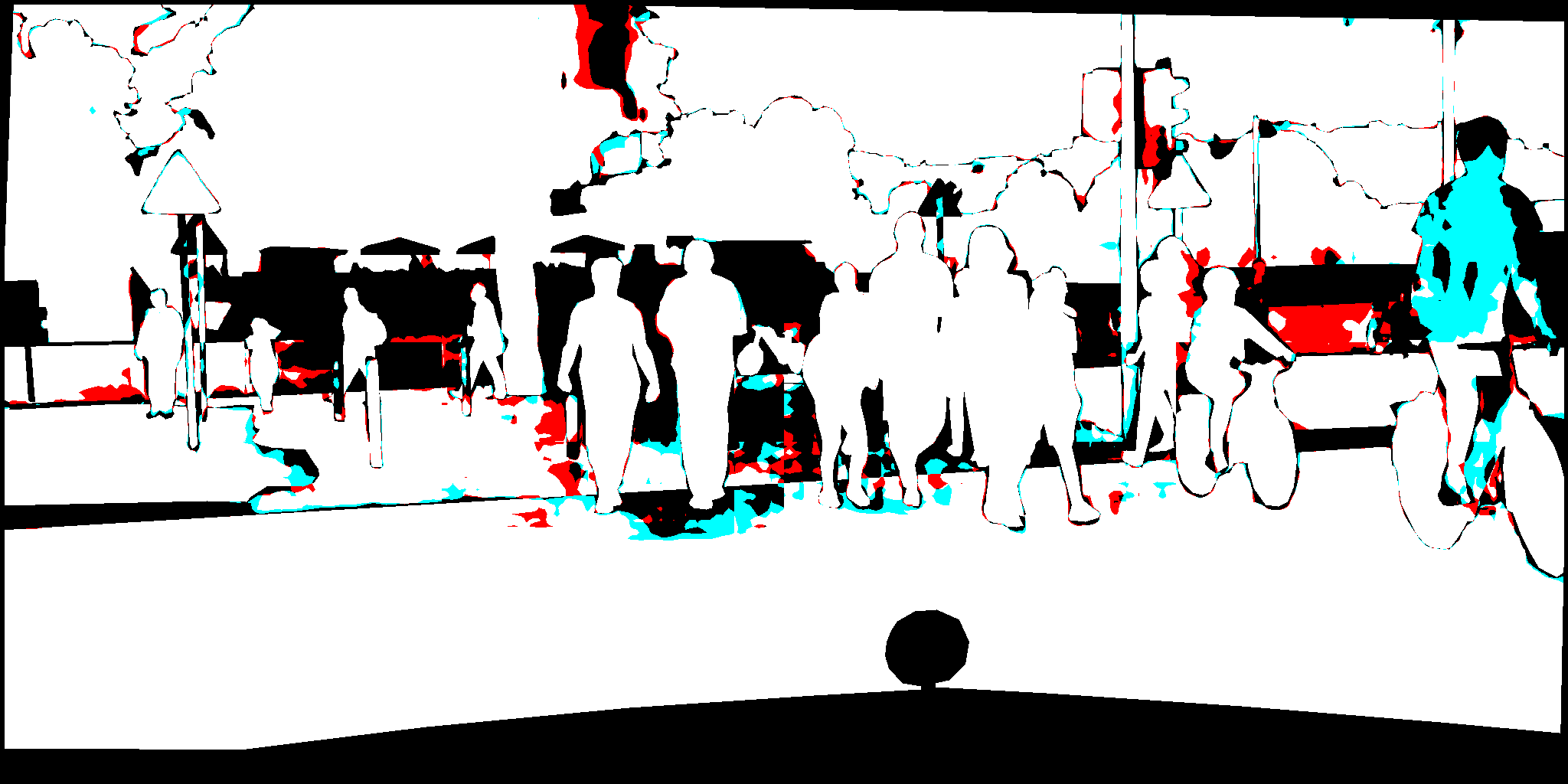}&
\includegraphics[width=0.22\linewidth]{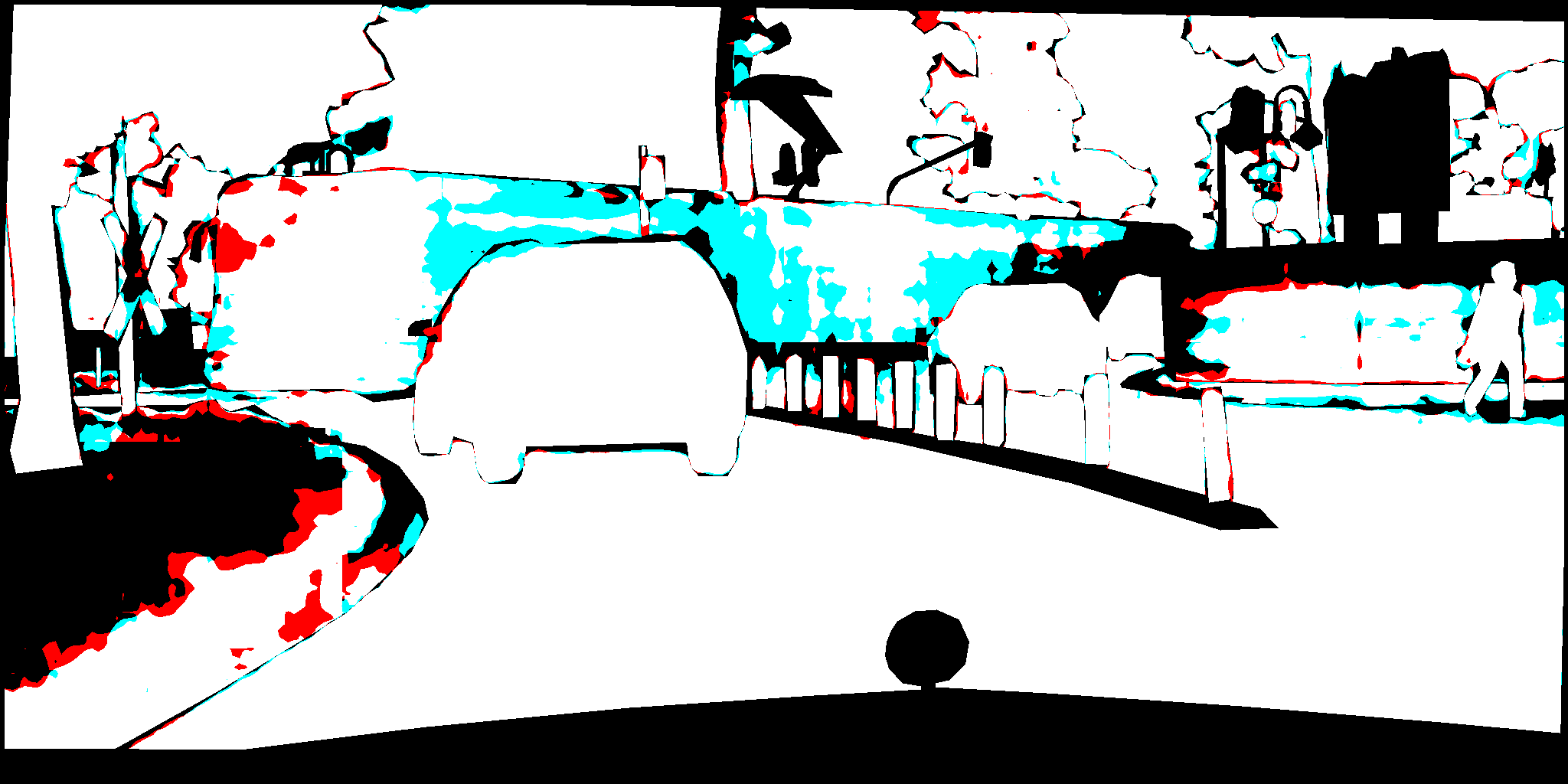}&
\includegraphics[width=0.22\linewidth]{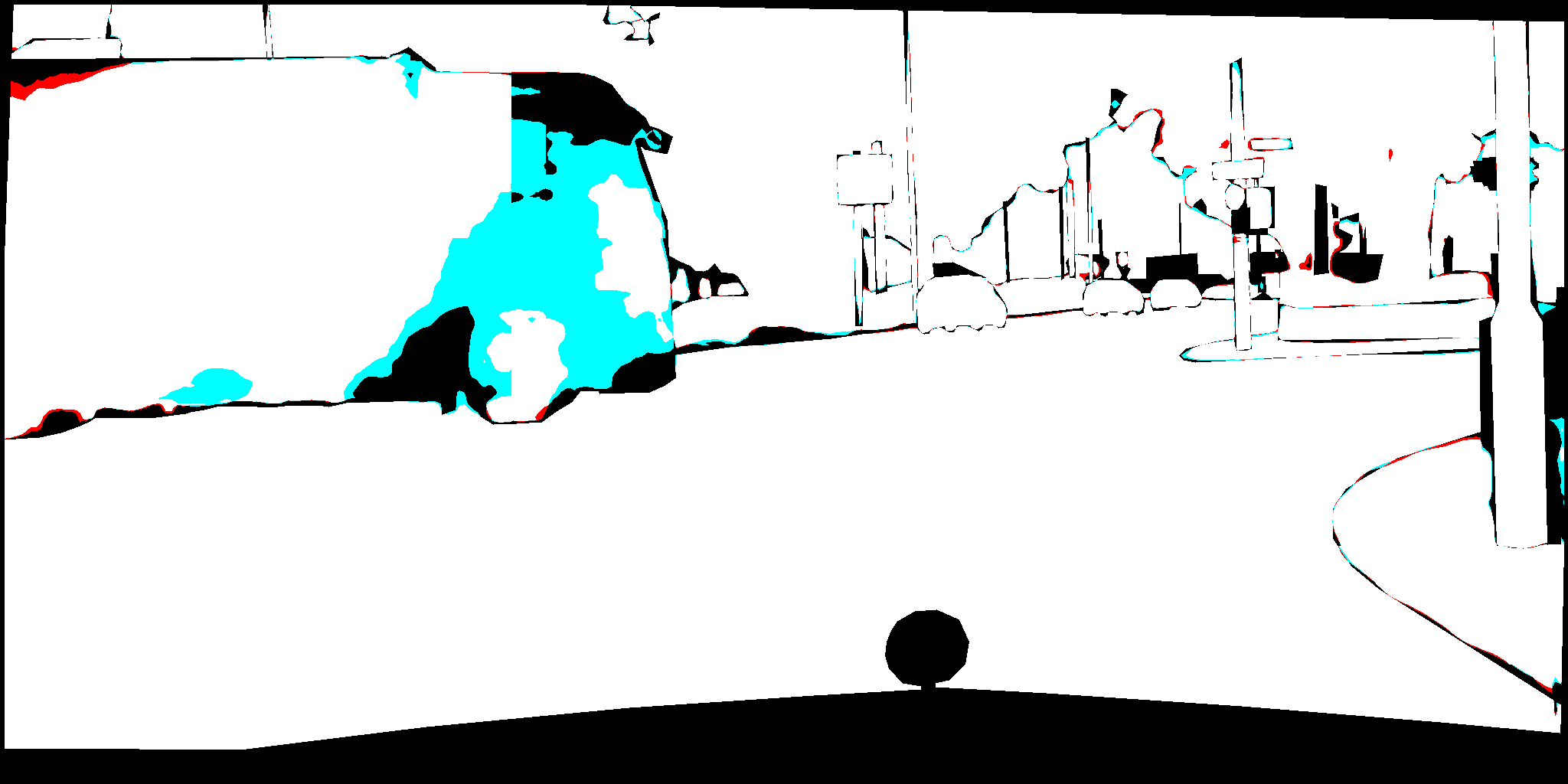}&
\includegraphics[width=0.22\linewidth]{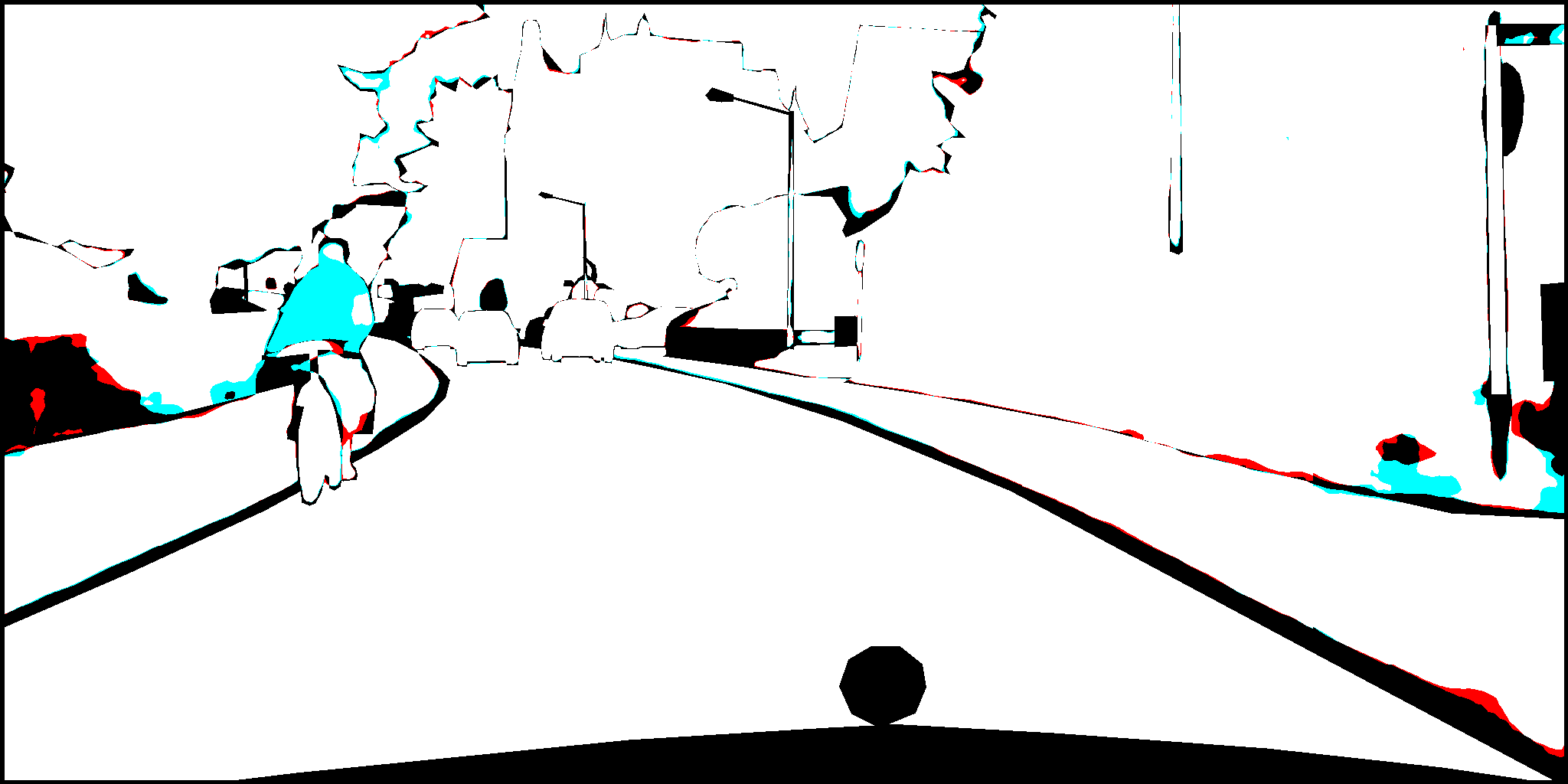}\\
\end{tabular}
}
\caption{Qualitative results on the Cityscapes validation set for Deeplab-v2 \cite{chen2016deeplab} (left) and PSPNet \cite{pspnet2016zhao} (right). The first four rows show the raw images, ground truth, baselines' predictions and our predictions. The last row is a visual comparison of correctly classified pixels. In white areas, both predictions are correct, in {\it red} areas, only the baseline prediction is correct and in {\it cyan} colored areas, the proposed predictions are correct, while the baseline prediction is  erroneous. 
}
\label{fig:visualization_cityscape}
\end{figure*}


\subsection{CamVid}
\label{subsec:camvid}
\begin{table}[t]
\tiny
  \begin{center}
    \label{table:camvid}
    \begin{tabular}{clccccccccccccccc}
      \toprule
       & Method &  & \rotatebox{75}{Building} & \rotatebox{75}{Tree} & \rotatebox{75}{Sky} & \rotatebox{75}{Car} & \rotatebox{75}{Sign} & \rotatebox{75}{Road} & \rotatebox{75}{Pedestrian} &  \rotatebox{75}{Fence}& \rotatebox{75}{Pole} & \rotatebox{75}{Sidewalk} & \rotatebox{75}{Bicyclist} & & \rotatebox{75}{\textit{mean IoU}} & \\
      \cmidrule(lr){1-17}
       &  Dilation \cite{YuKoltun2016}  & & 82.6 & 76.2 & 89.9 & 84.0 & 46.9 & 92.2 & 56.3 & 35.8 & 23.4 & 75.3 & 55.5  &  & 65.3 & \\
	   &   FSO \cite{kundu2016feature} &   &   84 &  77.2 & 91.3 & 85.6 & 49.9 & 92.5 & 59.1 & 37.6 & 16.9 & 76.0 & 57.2 &  & 66.1 & \\
     &   Synthetic \cite{richter2016playing} &   &  84.4 & 77.5 & 91.1 & 84.9 & 51.3 & 94.5 & 59  & 44.9 & 29.5 & 82  & 58.4 &  & 68.9 & \\
      \cmidrule(lr){1-17}
      &   Deeplab-v2 \cite{chen2016deeplab} &   &   83.8 &  \textbf{76.5} & 90.9 & \textbf{89.1} & 46.0 & \textbf{94.6} & 57.0 & 28.4 & 19.6 & \textbf{81.4} & 50.1 &  & 65.2 \\
      &   Deeplab-v2 + ours     &  &  \textbf{84.1} & 76.3   & 90.9 & 88.8 & \textbf{47.5} & 94.4 & \textbf{58.1} & \textbf{32.4} & \textbf{20.0} & 80.8 & \textbf{54.0} & & \textbf{66.1} &  \\
	  \cmidrule(lr){1-17}
      &   PSPNet \cite{pspnet2016zhao}      &     &  87.8 & \textbf{79.5}   & \textbf{91.4} & 91.4 & 57.7 & 96.5 &  66.7 & 58.6 & \textbf{23.5} & 87.8 & 66.9 & & 73.4 &  \\
      &   PSPNet + ours   &   & \textbf{88.0}  &  79.3  & 91.3 & \textbf{91.7} & \textbf{58.8} & 96.5 & \textbf{66.9} & \textbf{61.9} & 23.0 & 87.8 & \textbf{69.4} & & \textbf{73.9} &  \\
      \bottomrule
    \end{tabular}
  \end{center}
\caption{Quantitative results on CamVid dataset. With the proposed dilated convolutions, our method achieves better performance than two baselines, and we present new state-of-the-art performance on the CamVid dataset.}

\end{table}
CamVid is a smaller street view dataset captured from onboard camera. We not only compare our networks to baseline methods, but also compare to previous state-of-the-art methods on Camvid. For the ease of comparison, we utilized the training and test setup from \cite{sturgess2009combining}, which has 11 semantic classes, 367 training, 100 validation and 233 test images. The image resolution in our experiments is $640\times 480$.
The quantitative results are summarized in Tab.~4. We improve over Deeplab-v2 and PSPNet for 0.9 $pp$ and 0.5 $pp$, respectively. For most classes, we obtain comparable performance. Particularly, in the classes of ``Sign'', ``Fence'' and ``Bicyclist'', our method achieves clear improvements over Deeplab-v2 as well as PSPNet. This shows the benefit of our learned dilation: State-of-the art methods can be improved to recognize a range of classes better.



    \section{Conclusion}
    \label{sec:conclusion}
	In this paper, we have presented learnable dilated convolutions, which is fully compatible with existing architectures and adds only little overhead. We have applied our novel convolutional layer to learn channel-based dilation factors in the semantic segmentation scenario. Thus, we were able to improve the performance of Deeplab-LargeFOV, Deeplab-v2 and PSPNet for the semantic segmentation of street scenes consistently across two datasets. We showed that our method is able to obtain visually more convincing results, and improved quantitative performance. Besides, a series of ablation studies shows that learning the dilation parameter is helpful to design better semantic segmentation models in practice. 
    
	\bibliographystyle{splncs03}
	\bibliography{egbib}

\end{document}